\documentclass[10pt,twocolumn,letterpaper]{article}

\usepackage[pagenumbers]{cvpr} 
\usepackage{ulem}

\usepackage[dvipsnames]{xcolor}

\usepackage{comment}
\usepackage{tabularx}
\usepackage{multirow}

\newcolumntype{Y}{>{\centering\arraybackslash}X}

\usepackage[ruled,]{algorithm2e}

\newlength\mylen
\newcommand\myinput[1]{%
  \settowidth\mylen{\KwData{}}%
  \setlength\hangindent{\mylen}%
  \hspace*{\mylen}#1\\}
\SetKwComment{Comment}{$\triangleright$\ }{}

\DeclareMathOperator*{\argmin}{arg\,min}
\DeclareMathOperator*{\argmax}{arg\,max}

\definecolor{cvprblue}{rgb}{0.21,0.49,0.74}
\usepackage[pagebackref,breaklinks,colorlinks,citecolor=cvprblue]{hyperref}

\def\paperID{4502} 
\def\confName{CVPR}
\def\confYear{2024}

\title{Coupled Laplacian Eigenmaps for Locally-Aware 3D Rigid Point Cloud Matching}

\author{Matteo Bastico\textsuperscript{1}\footnotemark, Etienne Decenci\`ere\textsuperscript{2},
Laurent Cort\'e\textsuperscript{1}, Yannick Tillier\textsuperscript{3}, David Ryckelynck\textsuperscript{1}\\
Mines Paris, Université PSL\\
\textsuperscript{1}Centre des Matériaux (MAT), UMR7633 CNRS, 91003 Evry, France \\
\textsuperscript{2}Centre de Morphologie Mathématique (CMM), 77300 Fontainebleau, France \\
\textsuperscript{3}Centre de Mise en Forme des Matériaux (CEMEF), UMR7635 CNRS, 06904 Sophia Antipolis, France\\
}

\begin{document}
\maketitle
{\renewcommand*{\thefootnote}{\fnsymbol{footnote}}\stepcounter{footnote}%
  \footnotetext{Corresponding author: \href{mailto:matteo.bastico@minesparis.psl.eu}{matteo.bastico@minesparis.psl.eu}}}
\setcounter{footnote}{0}
\begin{abstract}
Point cloud matching, a crucial technique in computer vision, medical and robotics fields, is primarily concerned with finding correspondences between pairs of point clouds or voxels. In some practical scenarios, emphasizing local differences is crucial for accurately identifying a correct match, thereby enhancing the overall robustness and reliability of the matching process. Commonly used shape descriptors have several limitations and often fail to provide meaningful local insights about the paired geometries. In this work, we propose a new technique, based on graph Laplacian eigenmaps, to match point clouds by taking into account fine local structures. To deal with the order and sign ambiguity of Laplacian eigenmaps, we introduce a new operator, called Coupled Laplacian\footnote{Code: \url{https://github.com/matteo-bastico/CoupLap}}, that allows to easily generate aligned eigenspaces for multiple registered geometries. We show that the similarity between those aligned high-dimensional spaces provides a locally meaningful score to match shapes. We firstly evaluate the performance of the proposed technique in a point-wise manner, focusing on the task of object anomaly localization on the MVTec 3D-AD dataset. Additionally, we define a new medical task, called automatic Bone Side Estimation (BSE), which we address through a global similarity score derived from coupled eigenspaces. In order to test it, we propose a benchmark collecting bone surface structures from various public datasets. Our matching technique, based on Coupled Laplacian, outperforms other methods by reaching an impressive accuracy on both tasks. 
\end{abstract}
    
\section{Introduction}
\label{sec:intro}

Point cloud matching, or more generally 3D shape matching, is a fundamental task in computer vision. It involves finding the closest matching geometry to a target shape within a set of reference shapes \cite{tangelder_survey_2008}. 
In addition, if the task involves finding rigid transformations that best align the target shape with the reference, it is often part of a registration process. In particular, point-set \textit{rigid registration} determines the relative transformation needed to align two point clouds without altering their internal structures \cite{maiseli_recent_2017}. This problem is essential for many practical computer vision tasks, such as medical image analysis \cite{baum_real-time_2021, sinko_3d_2018, kobayashi_sketch-based_2023, pilevar_cbmir_2011}, intelligent vehicles \cite{gao_shrec_2023, li_rigid_2018}, human pose estimation \cite{ge_articulated_2015} and objects retrieval and tracking \cite{tang_track_2022, nguyen_robot_2020}. Traditional \cite{besl_method_1992, fischler_random_1981, rusu_fast_2009} and probabilistic registration and matching methods \cite{eckart_fast_2018, gao_filterreg_2019, jian_robust_2011, myronenko_point-set_2010}, while robust, often struggle to optimally align complex geometries, especially in cases with intricate local structures or slight deformations. 

Over the years, several methods have been proposed to tackle the challenge of accurate and efficient 3D shape matching and retrieval \cite{bickel_novel_2023, belongie_shape_2002, bronstein_scale-invariant_2010, reuter_discrete_2009, reuter_laplace-spectra_2005, tangelder_survey_2008, xu_shape_2016}. Data-driven 3D shape descriptors \cite{rostami_survey_2019}, capturing underlying properties of the shapes under study, are the common denominator of early shape matching techniques. Global descriptors, such as volume and areas descriptors \cite{zhang_efficient_2001}, describe the entirety of the shape in one compact representation, often failing to capture local fine details of complex geometries. On the other hand, local descriptors \cite{lowe_distinctive_2004, rusu_fast_2009} aim to tackle this issue but they generally are sensitive to noise, based on landmarks and they might not capture semantic information \cite{tang_evaluation_2012}. More recently, deep-learned shape descriptors \cite{xie_deepshape_2017, bergmann_anomaly_2023} and neural networks for shape matching, based on auto-encoders \cite{xu_shape_2016} or transformers \cite{shajahan_point_2021, wang_deep_2019}, have been proposed. Despite their good performances, these methods require a huge amount of annotated data for training, which are hard to collect in fields such as medical imaging \cite{lotan_medical_2020}. Furthermore, non-rigid point cloud matching and retrieval methods \cite{lian_comparison_2013, lian_non-rigid_2015, wu_learning_2023} are designed to handle shape deformations and, therefore, they might be excessively flexible ignoring fine local details that are not due to deformations, such as anomalies. 

In this study, we introduce a novel method for rigid 3D point cloud matching, based on spectral Laplacian eigenmaps \cite{belkin_laplacian_2003, ghojogh_laplacian-based_2022}, which focus on local details. This technique is designed to overcome the limitations of shape descriptors and, although it can be seen as a linear graph neural network, it does not require training. One of the main reasons that combinatorial and geometric Laplacians are often considered for spectral graph processing is that their eigenvectors possess properties like the classical Fourier basis functions \cite{zhang_spectral_2010}. We leverage these characteristics to perform locally-aware shape comparison of point clouds, equipped with k-nearest neighbor graph, without relying on additional descriptors. Nevertheless, eigenspaces alignment, including both eigenvalue ordering and sign disambiguity of the eigendecomposition, is required to correctly match shapes in the spectral space. Current State-of-The-Art methods for such alignment are based on Laplacian eigenfunctions matching \cite{mateus_articulated_2008, sharma_3d_2021}, i.e. matching the histograms of the eigenvectors. The latter is not robust and frequently fails in possible scenarios with highly symmetric geometries, such as bones, {with small eigenvalue separation. In this work, we introduce a novel operator, called \textit{Coupled Laplacian}, designed to simultaneously produce aligned eigenspaces for multiple registered geometries. Namely, we show that when two or more shapes, each with their corresponding graphs, are merged into a single graph using artificial cross-edges, the eigendecompositon of the Laplacian derived from such a combined graph yields aligned spectral spaces for each individual component. Furthermore, this method is order-invariant, count-agnostic and landmarks-free, enabling it to handle point clouds without being influenced by their specific arrangement or the number of points contained. Finally, we utilize the distance between these aligned higher-dimensional spectral spaces, which accounts for intricate local structures, as a global or point-wise score for different applications of 3D shape matching, as shown in \cref{fig:intro}. 

\begin{figure}[t]
    \centering
    \includegraphics[width=0.4\textwidth]{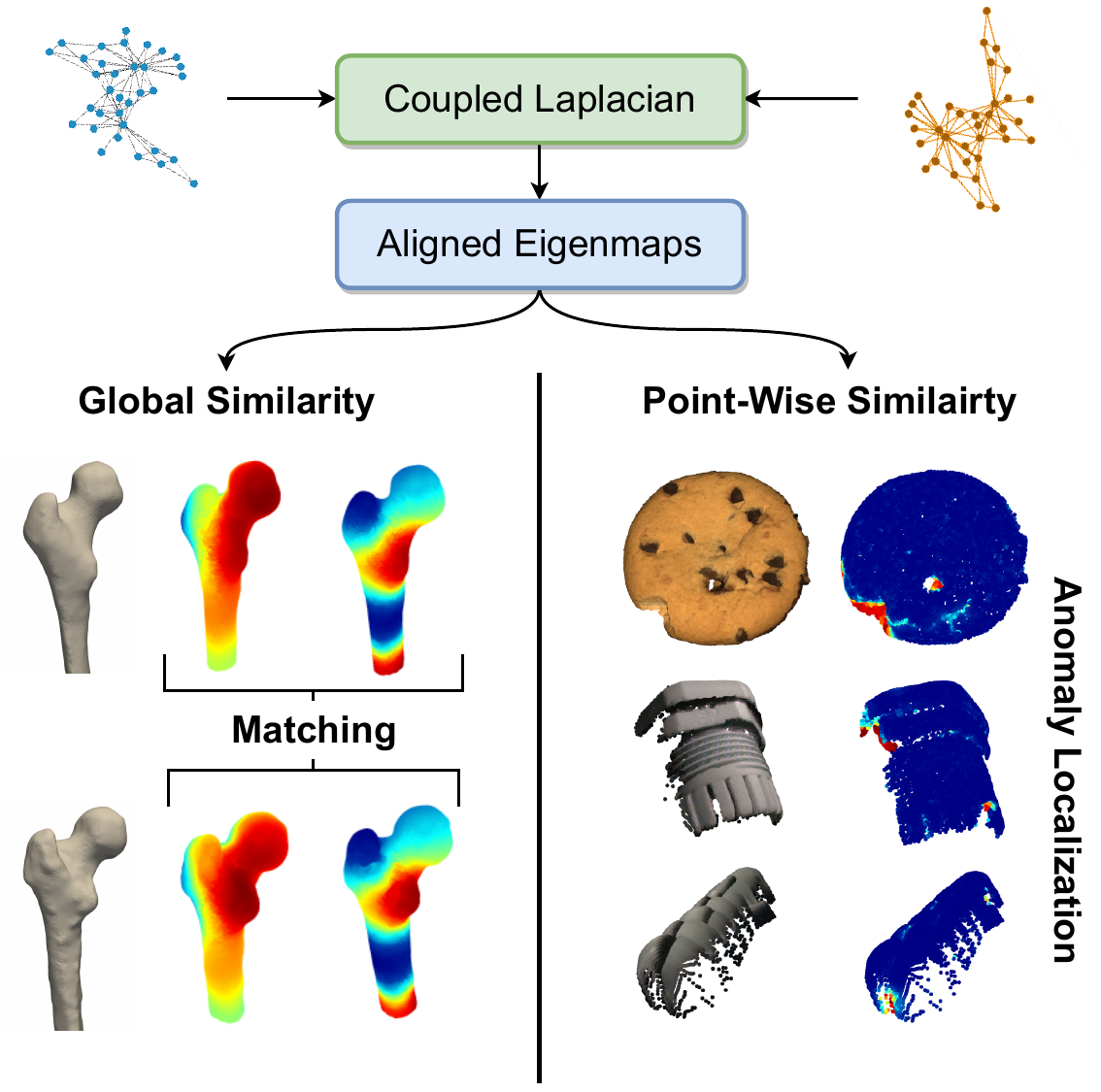}
    \caption{Overview of the proposed technique applied to different tasks. The global similarity between aligned eigenmaps of multiple geometries, generated from the Coupled Laplacian, is used to match bones and predict their body side. While local similarity is exploited for accurate 3D anomaly detection.}
    \label{fig:intro}
\end{figure}

One natural application of the proposed technique, is 3D object anomaly detection thought the identification of local differences between two shapes. Hence, we tested our method on the MVTec 3D-AD dataset \cite{bergmann_mvtec_2022}, recently proposed for unsupervised point cloud anomaly detection and localization. Furthermore, in the context of medical imaging, correctly identifying the side of a bone (left or right) is crucial for diagnosis, treatment planning, or skeletal analysis \cite{corballis_bilaterally_2020}. We refer to this task as \textit{Bone Side Estimation} (BSE). Recent studies on bone landmark detection \cite{fischer_robust_2020} and statistical shape modeling \cite{keast_geometric_2023} have highlighted the limitations of manual side identification for ensuring proper functioning. To the best of our knowledge, manual bone markings \cite{bandovic_anatomy_2023} are currently the only technique used for this purpose. Consequently, automatic BSE arises as an interesting, yet challenging, task to assist the development of fully automated pipelines for medical image analysis, such as patient-specific preoperative planning for Total Knee Arthroplasty (TKA) \cite{lambrechts_artificial_2022} or Anterior Cruciate Ligament Reconstruction (ACLR) \cite{figueroa_navigation_2023, morita_computer-aided_2015}. Due to the bilateral symmetry of the animal and human body \cite{hollo_manoeuvrability_2012, toxvaerd_emergence_2021}, i.e. right and left sides are mirror shapes of one another, the BSE can be defined as a non-trivial chiral shape matching problem. 
Then, given a known reference or source bone, assessing the side of a target bone involves finding the best match between the source-target pair and its mirrored counterpart. The complexity of this task arises from subtle local differences in mirrored bones, which makes it a suitable application of the proposed shape matching technique utilizing coupled Laplacian eigenmaps as local geometry descriptors. We propose a benchmark for human BSE by extracting surface point clouds of different bones, e.g. femur, hip and tibia, from public datasets \cite{fischer_robust_2020, keast_geometric_2023, nolte_non-linear_2016}. 
We also discuss, in the Supplementary Material, a non-rigid correspondence application of the proposed matching algorithm, the cross-species BSE (human to animal), on an internal dataset \cite{maeztu_redin_wear_2022}. 

Our contributions are summarized as follows:

\begin{itemize}[noitemsep,topsep=0pt]
    \item We propose a new method to preform locally-aware 3D rigid point cloud matching, considering fine local structures, based on graph Laplacian eigenmaps.
    \item We define the Coupled Laplacian operator for aligned graphs to tackle the order and sign ambiguity issue of the eigendecomposition.
    \item Based on the proposed technique, we introduce a new method for automatic BSE to assist fully automated medical pipelines and we propose a benchmark to test it.
    \item We extensively evaluate our method on two tasks: anomaly detection and BSE, outperforming previously proposed techniques. 
\end{itemize}
\section{Related Works}
\label{sec:related_works}

Early efforts in developing hand-crafted 3D local features for point cloud matching, registration and retrieval have typically drawn inspiration from 2D descriptors. Several methods, such as Signature of Histograms of OrienTations (SHOT) \cite{salti_shot_2014, tombari_unique_2010}, Rotational Projection Statistics (RoSP) \cite{guo_rotational_2013} and Unique Shape Context (USC) \cite{tombari_unique_2010} rely on the estimation of an unique Local Reference Frame (LRF). The latter is usually based on the eigendecomposition of the covariance matrix of the neighbours of a point of interest, which are then projected into the LRF to analyze their geometric properties. For instance, SHOT \cite{salti_shot_2014} captures local shape and orientation information by computing histograms of surface normals and point distribution around keypoints. In contrast, LRF-free approaches \cite{birdal_point_2015, rusu_aligning_2008, rusu_fast_2009} try to rely just on features that are intrinsically invariant. The most common LRF-free features are Fast Point Feature Hisograms (FPFH) \cite{rusu_fast_2009} which, similarly to SHOT, calculates histograms of surface normals considering their relationships in a local region around a keypoint and generating $33$-dimensional descriptors. Despite the progress made with hand-crafted 3D local features, they encounter difficulties when it comes to deal with issues like point cloud resolutions, noisy data, and occlusions \cite{guo_comprehensive_2016}.

On the contrary, spectral-based point cloud descriptors \cite{aubry_wave_2011, hu_salient_2009, reuter_laplacebeltrami_2006,weinmann_semantic_2014, sun_concise_2009, rustamov_laplace-beltrami_2007, wang_robust_2019} are a category of feature extraction methods that leverage spectral analysis techniques from graph theory to capture the underlying structure and intrinsic geometric properties of point clouds. Among them, shape-DNA \cite{reuter_laplacebeltrami_2006} is a surface descriptor based on eigenvalue analysis of the Laplace-Beltrami operator. They propose to use the sequence of eigenvalues (spectrum) of the Laplace operator as a fingerprint characterizing the intrinsic geometry of 3D shapes represented as point clouds. This method has been used for shape retrieval, classification, and correspondence. Weinmann \textit{et al.} \cite{weinmann_semantic_2014} also proposed to extract a features set consisting of $8$ eigenvalues-based indices for each 3D point of a cloud. Furthermore, Heat Kernel Signature (HKS) \cite{sun_concise_2009} and Wave Kernel Signatures (WKS) \cite{aubry_wave_2011} are descriptors measuring how heat and wave propagate across a shape, having the eigendecomposition as leading element of the computation. Scaled eigenvectors evaluated at each point are instead directly exploited by the Global Point Signature (GPS) \cite{rustamov_laplace-beltrami_2007} to represent a point cloud as a set of infinite-dimensional vectors, characterizing each point within the global context of the surface it belongs. Nevertheless, the vast majority of these works assumes that the eigenvalues of a shape are distinct and, therefore, can be ordered. Indeed, in practice, due to numerical approximations, we cannot guarantee that the eigenvalues of the Laplacian are all distinct and, possible symmetries in the shapes may cause some of them to have multiplicity grater than one. As shown by Mateus \textit{et al.} \cite{mateus_articulated_2008}, when dealing with shape matching, an elegant way to overcome this problem is to use the Laplacian eigenmaps scheme \cite{belkin_laplacian_2003}, which can be seen as a reduced GPS, and perform a-posteriori alignment of the resulting point cloud embeddings. Matching the eigenfunctions histograms, i.e. their signatures, is the only reliable method for such embeddings alignment \cite{mateus_articulated_2008, sharma_3d_2021}. Recently, \textit{Ma et. al.} \cite{ma_laplacian_2024} proposed a canonization algorithm for sign and basis invariance, called Maximal Axis Projection (MAP), that adopts the permutation-invariant axis projection functions to determine the canonical directions. Unfortunately, when the eigenvalues separation is small due to symmetries in the geometries, these methods becomes unstable and sensitive to noise. To overcome this issue, in this work we propose the Coupled Laplacian operator to produce a-priori aligned eigenmaps for several registered shapes, and perform locally-aware shape matching.

Deep-learned point-cloud descriptors \cite{gojcic_perfect_2019, xie_deepshape_2017, bergmann_anomaly_2023, zeng_3dmatch_2017} arose as an alternative to generate local features for 3D sufraces. We can distinguish three main categories depending on the backbone architecture employed. Convolutional Neural Networks (CNNs) are often used on point clouds projected into 2D depth images \cite{elbaz_3d_2017, huang_learning_2017} or directly on 3D voxels \cite{gojcic_perfect_2019, zeng_3dmatch_2017, huang_learning_2017}. 
Secondly, to work directly on raw point cloud data, PointNet and PointNet++ have been proposed aiming to learn rotation and permutation invariant features \cite{qi_pointnet_2017, qi_pointnet2_2017}. Based on PointNet, several learned descriptors have been introduced \cite{deng_ppfnet_2018, deng_ppf-foldnet_2018, yew_3dfeat-net_2018, yi_lift_2016}. Among them, PPFNet \cite{deng_ppfnet_2018} and PPF-FoldNet \cite{deng_ppf-foldnet_2018} try to improve the feature representation of PointNet by incorporating global context and point-pair features. Nevertheless, the lack of convolutional layers in these models limits the learning of local geometries. Finally, transformers-based descriptors, such as Deep Closest Point \cite{wang_deep_2019}, have been recently proposed trying to exploit the attention mechanisms to capture shapes and surfaces intrinsic characteristics. Arguably, training a deep-learning model, especially if based on transformers, it is not always feasible in terms of training samples required to achieve good performances. Furthermore, in supervised algorithms \cite{zeng_3dmatch_2017, gojcic_perfect_2019, deng_ppfnet_2018, deng_ppf-foldnet_2018}, paired 3D patches, such as in the 3DMatch dataset \cite{zeng_3dmatch_2017}, are needed to train the models. For this reason, the 3D rigid point cloud matching method we propose does not need training and utilizes only the similarity among properly aligned spectral embeddings to relate an unseen target shape to the given references. 
\section{Method}
\label{sec:method}

\begin{figure*}
    \centering
    \includegraphics[width=0.79\textwidth]{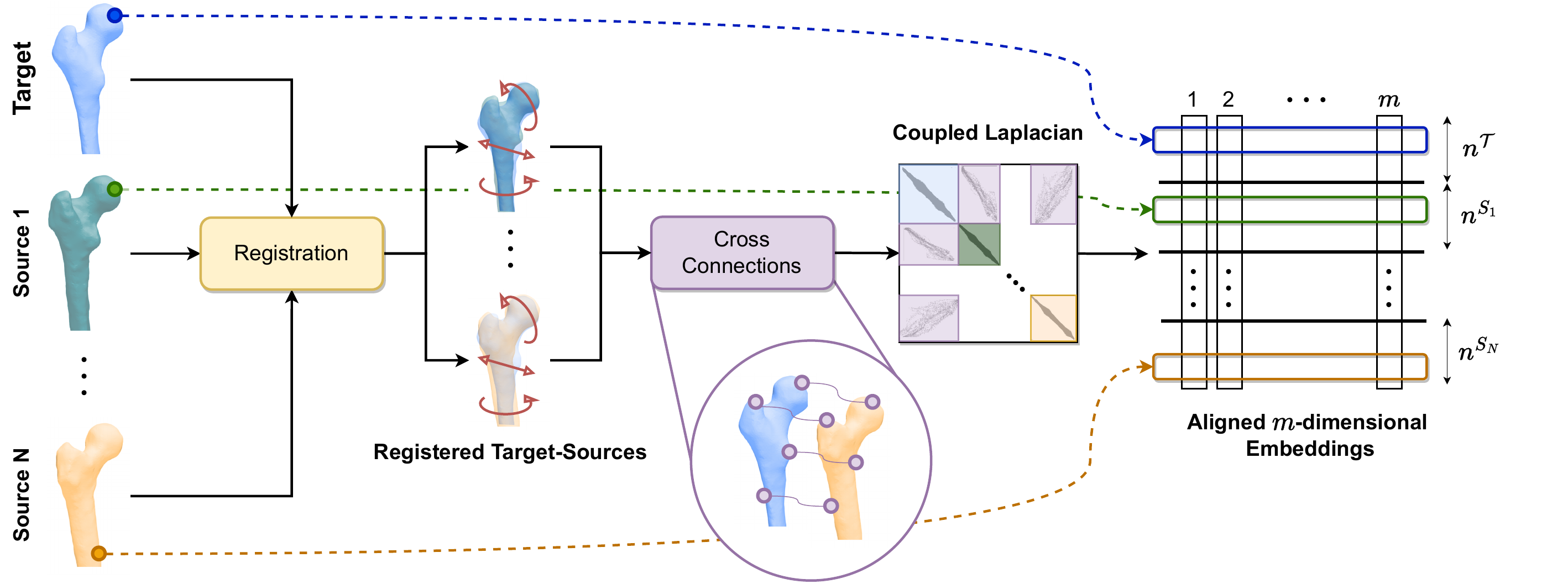}
    \caption{Proposed workflow of the Coupled Laplacian applied to proximal femur shapes. The $N$ sources are registered to the target using a rigid or affine registration. After that, cross-connection are added between each target-source pair (for simplicity, in the zoom the shapes are not overlapping) and the Coupled Laplacian is computed on the global graph. Its eigendecomposition leads to aligned spectral embeddings of the input geometries that can be used for shape matching.}
    \label{fig:coupled-laplacian}
\end{figure*}
\textbf{Graph Laplacian.} A 3D point-cloud $\{\pmb{x}_i\}_{i=1}^n$ can be treated as a connected undirected weighted graph $\mathcal{G}(\mathcal{V}, \mathcal{E})$, where $\mathcal{V} = \{\pmb{x}_i\}_{i=1}^n$ is the nodes set and $\mathcal{E} = \{\pmb{e}_{ij}\}$ is the edge set. The latter is generally obtained though the construction of a nearest neighbors graph \cite{belkin_laplacian_2003}. To that purpose, in this work, we consider the $k$-Nearest Neighbour ($k$-NN) approach, that is, node $j$ is connected to node $i$ if $\pmb{x}_i$ is among the $k$ nearest neighbours of $\pmb{x}_j$. Hence, we can build a weighted adjacency matrix, $\pmb{W} = \{ w_{ij} \}$, which stores the connections between nodes. In particular, in spectral graph theory, a Radial Basis Function (RBF) is commonly used as weight function for the edges between $\pmb{x}_i$ and $\pmb{x}_j$, as 
\begin{equation}
\label{eq:RBF}
    w_{ij} = \exp \left( - \frac{\text{d}^2(\pmb{x}_i, \pmb{x}_j)}{\sigma^2} \right)
\end{equation}
where $\text{d}^2(\cdot, \cdot)$ is the squared Euclidean distance between two vertices and $\sigma^2$ a free parameter which, for simplicity, we set to the maximum squared distance between connected nodes, $\max_{ij}\text{d}^2(\pmb{x}_i, \pmb{x}_j)$. The \textit{Laplacian matrix}, $\textbf{L} \in \mathbb{R}^{n \times n}$, of a graph constructed in such way, is defined as $\textbf{L} = \textbf{D} - \textbf{W}$
where $\mathbb{R}^{n \times n} \ni \pmb{D} = \text{diag}([d_1, \cdots, d_n$]) is the degree matrix with diagonal elements $d_i = \sum_{j=1}^n w_{ij}$ \cite{merris_laplacian_1994}. The Laplacian eigenvalues, $\{\lambda_i\}_{i=0}^n$, and eigenvectors, $\{\pmb{\phi}_i\}_{i=0}^n$, can be computed by solving the generalized eigenproblem
\begin{equation}
    \label{eq:eigenproblem}
    \textbf{L}\pmb{\phi}_i = \lambda_i\pmb{B}\pmb{\phi}_i, \;\; i=0,\cdots, n.
\end{equation}
where $\pmb{B} \in \mathbb{R}^{n \times n}$ is generally set as $\pmb{D}$, and $\lambda_i \le \lambda_{i+1}$. Finally, eigenmaps are simply eigenvectors sub-spaces, generated by leaving out $\pmb{\phi}_0$, corresponding to $\lambda_0 = 0$, and using the next $m$ eigenvectors for embedding graph nodes in an $m$-dimensional space, $\pmb{x}_i \rightarrow [\pmb{\phi}_1(i), \cdots, \pmb{\phi}_m(i)]$

\textbf{Coupled Laplacian.} The complete pipeline to generate aligned Laplacian embeddings through graph coupling is illustrated in \cref{fig:coupled-laplacian}. Let $\mathcal{G}^\mathcal{T}$ be the graph of a target 3D point cloud and $\mathcal{G}^{S_1}, \cdots, \mathcal{G}^{S_N}$ the ones of $N$ different sources. We construct a global graph, or coupled graph, $\mathcal{G}^{C}$, by adding \textit{cross-connections} separately between the vertices of the source, $\mathcal{V}^\mathcal{T} = \{\pmb{x}^\mathcal{T}_i\}_{i=1}^{n^\mathcal{T}}$, and each of the reference shapes, $\mathcal{V}^{S_k}= \{\pmb{x}^{S_k}_i\}_{i=1}^{n^{S_k}}$ with $k=1, \cdots, N$. To include meaningful cross-connections, we first perform a rigid \cite{fischler_random_1981, gao_filterreg_2019, myronenko_point-set_2010} or affine \cite{myronenko_point-set_2010} registration of the source geometries to align to the target. These methods are preferred to a non-rigid registration because, when dealing with the identification of fine local variations, they do not change the relative position of the points inside a point-set and, therefore, the deformations are kept unchanged. After that, a sub-set of vertices, $F^\mathcal{T} \subset \mathcal{V}^\mathcal{T}$ , is stochastically extracted from the target geometry and their nearest correspondences are searched in each of the aligned reference point clouds as
\begin{equation}
    \label{eq:nearest}
    F^{S_k} = \{\pmb{f}^{S_k}_{j}:\argmin_{\pmb{x}^{S_k}_i \in \mathcal{V}^{S_k}} \text{d}^2(\pmb{x}^{S_k}_i, \pmb{f}^\mathcal{T}_j) | \pmb{f}^\mathcal{T}_j \in F^\mathcal{T}\}
\end{equation}
with $k=1, \cdots, N$. In this way, there is no constraint on the original number points comprising each shape nor on their initial coordinates systems. Furthermore, the cardinality of the target sub-set, is chosen as a fraction of its total points, $|F^\mathcal{T}| = l \cdot n^\mathcal{T}$, where $0 < l \le 1$. The cross-connection are therefore added for each pair $(\pmb{f}^{S_k}_i, \pmb{f}^\mathcal{T}_i)$, $1\le i \le |F^\mathcal{T}|$, between the target and the $k$-th source. Finally, we arrange the vertex indices of the coupled graph such that they are grouped for each individual geometry, and the whole weighted adjacency matrix can be computed as in \cref{eq:RBF}.

We define the Laplacian matrix of a global graph, constructed as describe above, as the Coupled Laplacian, $\pmb{L}^C \in \mathbb{R}^{n \times n}$, where $n = n^\mathcal{T} + \sum_{k=1}^N n^{S_k}$.  Thanks to the vertex ordering of the coupled graph, the solution of the generalized eigenproblem of \cref{eq:eigenproblem} applied to the Coupled Laplacian, with $\pmb{B}^C =\text{diag}([\pmb{D}^\mathcal{T}, \pmb{D}^{S_1}, \cdots, \pmb{D}^{S_N}])$
 , yields to \textit{coupled eigenvectors} $\{\pmb{\phi}^C_i\}_{i=0}^n$. These eigenvectors can then be split into each single component of the coupled graph as 
\begin{equation}
\label{eq:split}
    \pmb{\phi}^C_i = [\pmb{\phi}^\mathcal{T}_i, \; \pmb{\phi}^{S_1}_i, \; \cdots, \;\pmb{\phi}^{S_N}_i]^T.
\end{equation}

The eigenspaces restricted to the single geometries, obtained thought the Coupled Laplacian, are intrinsically aligned up to a certain component, depending of the factor $l$, and the eigenvalues ordering issue is automatically solved since the split eigenvectors are associated to the same eigenvalues. The proof of this property for the ideal case of perfect match and more theoretical aspects of the Coupled Laplacian are reported in the Supplementary Material.

\textbf{Shape Matching with Eigenmaps.} The Coupled Laplacian allows the generation of $m$-dimensional embeddings aligned for the graph vertices of the target and reference shapes, $\pmb{\Phi}^\mathcal{T} = (\pmb{\phi}^\mathcal{T}_1, \cdots, \pmb{\phi}^\mathcal{T}_m) \in \mathbb{R}^{n_T \times m}$ and $\pmb{\Phi}^{S_k} =  (\pmb{\phi}^{S_k}_1, \cdots, \pmb{\phi}^{S_k}_m) \in \mathbb{R}^{n_{S_k} \times m}$ for $1 \le k \le N$, respectively. Therefore, a comparison of multiple geometries in this higher dimensional spectral space yields to a proper consideration of local structures. We consider in the following two possible applications of the coupled eigenmaps: (1) a global shape matching score and (2) local similarity scores. The first one can be obtained by measuring the similarity between the aligned eigenspaces through the Grassmann distance, $d_G(\cdot, \cdot)$ \cite{ye_schubert_2016}. Nevertheless, it cannot be computed directly, since we consider point clouds of arbitrary size and we do not have point-to-point correspondences. Hence, we restrict the distance computation only to the set of cross-connected vertices, $F^\mathcal{T}$ for the target and $F^{S_k}$ for the $k$-th reference. In order to get the reduced basis restricted only to those coupled points, we perform a QR decomposition of the restriction of the eigenmodes to the cross-points. The best matching source is then given by 
\begin{equation}
    \label{eq:grassman}
    \argmin_k d_G(\pmb{Q}^\mathcal{T}, \pmb{Q}^{S_k}).
\end{equation}
where $\pmb{Q}^\mathcal{T}$ and $\pmb{Q}^{S_k}$ are obtained from the QR factorization of $\pmb{\Phi}^\mathcal{T}(F^\mathcal{T}, :)$ and $\pmb{\Phi}^{S_k}(F^{S_k}, :)$, respectively.
This first approach is used in the following to solve the BSE task by using two source shapes, i.e. the reference bone of known side and its contralateral mirrored version. 

On the other hand, to obtain point-wise similarity scores we propose to compare the $m$-dimensional embeddings of the cross-connection vertices by using the cosine distance function. In this case, we interpret the distance value as the probability of local structural difference, where 0 means that the two compared points have the same local structure. 

\section{Experiments}

\subsection{Experiment settings}
\label{sec:settings}

\begin{table}[t]
\footnotesize
\caption{Summary of the point cloud bone structures collected for the BSE benchmark. L and R stands for Left and Right, respectively, and \textbf{S} for Sheep.}
    \centering
\begin{tabularx}{0.45\textwidth}{Yl|YY|YY|YY|YY}
\toprule
 &\multirow{2}{*}{Dataset} & \multicolumn{2}{c|}{Femur} & \multicolumn{2}{c|}{Hip} & \multicolumn{2}{c|}{Tibia} & \multicolumn{2}{c}{Fibula}\\
 \cline{3-10}
 & & L & R & L & R & L & R & L & R \\
 \midrule
 \multirow{3}{*}{\rotatebox[origin=c]{90}{\textbf{Human}}} & \textit{Fisher et al.} \cite{fischer_robust_2020} & 18 & 19 & 20 & 20 & - & - & - & -\\
  
  & SSM-Tibia \cite{keast_geometric_2023} & - & - & - & - & - & 30 & - & 30\\ 
  
  & ICL \cite{nolte_non-linear_2016} & 35 & 35 & - & - & 35 & 35 & 35 & 35\\
  \midrule
  \rotatebox[origin=c]{90}{\textbf{S}}& Internal \cite{maeztu_redin_wear_2022} & 18 & 18 & - & - & 18 & 18 & - & - \\
  \midrule
  & Total & 71 & 72 & 20 & 20 & 53 & 83 & 35 & 65 \\
\bottomrule
\end{tabularx}
    \label{tab:benchmark}
\end{table}
\begin{table*}[t]
\footnotesize
\caption{Anomaly localization results. The area under the PRO curve is reported for an integration limit of $0.3$ for each
evaluated method and dataset category. GAN, AE and VM results are provided by \textit{Bergmann et al.} \cite{bergmann_mvtec_2022}. Moreover, we include the results obtained by restricted GPS \cite{rustamov_laplace-beltrami_2007}, eigenfunctions (Hist) \cite{mateus_articulated_2008, sharma_3d_2021} and Euclidean matching and MAP \cite{ma_laplacian_2024}. All the matching methods are applied after affine CPD registration, when not specified, or CMM + CPD Non-Rigid (\textit{NR}) registration. The subsctipt on a method indicates the number of eigenmaps used. The overall best performing methods are highlighted in boldface, while the bests for each category are underlined.}
    \centering
\begin{tabularx}{\textwidth}{Yl|YYYYYYYYYY|l}
\toprule
  & Method & Bagel & Cable Gland & Carrot & Cookie & Dowel & Foam & Peach & Potato & Rope & Tire & Mean $\uparrow$\\
 \midrule
 \multirow{3}{*}{\rotatebox[origin=c]{90}{\textbf{3D RGB}}} & VM & 0.388 & 0.321 & 0.194 & 0.570 & 0.408 & 0.282 & 0.244 & 0.349 & 0.268 & 0.331 & 0.335\\
 & GAN & 0.421 & \underline{0.422} & 0.778 & \underline{0.696} &  0.494 &  0.252 & \underline{0.285} & \underline{0.362} & 0.402 & \underline{0.631} & 0.474\\
  & AE & \underline{0.432} & 0.158 & \underline{\textbf{0.808}} & 0.491 & \underline{\textbf{0.841}} & \underline{0.406} & 0.262 & 0.216 & \underline{0.716} & 0.478 & \underline{0.481} \\
 \midrule
 \midrule
 \multirow{11}{*}{\rotatebox[origin=c]{90}{\textbf{3D Only}}} & GAN & 0.111 & 0.072 & 0.212 & 0.174 & 0.160 & 0.128 & 0.003 & 0.042 & 0.446 & 0.075 & 0.143\\
  & AE & 0.147 & 0.069 & 0.293 & 0.217 & 0.207 & 0.181 & 0.164 & 0.066 & 0.545 & 0.142 & 0.203\\
  & VM & 0.280 & 0.374 & 0.243 & 0.526 & 0.485 & 0.314 & 0.199 & 0.388 & 0.543 & 0.385 & 0.374\\
  & Euclidean (\textit{NR}) & 0.404 & 0.623 & 0.731 & 0.366 & 0.771 & 0.303 & 0.590 & 0.772 & 0.697 & 0.583 & 0.584\\
  &  GPS$_{100}$\cite{rustamov_laplace-beltrami_2007} & 0.452 & 0.616 & 0.695 & 0.364 & 0.738 & 0.471 & 0.659 & 0.844 & 0.647 & 0.651 & 0.613\\
   & GPS$_{200}$\cite{rustamov_laplace-beltrami_2007} & 0.465 & 0.621 & 0.690 & 0.363 & 0.739 & 0.480 & 0.672 & 0.833 & 0.653 & 0.670 & 0.619\\
  & Hist$_{100}$ \cite{sharma_3d_2021}& 0.476 & 0.629 & 0.703 & 0.365 & 0.744 & 0.473 & 0.661 & 0.840 & 0.647 & 0.693 & 0.623\\
  & MAP$_{200}$ \cite{ma_laplacian_2024} & 0.481 & 0.630 & 0.694 & 0.399 & 0.742 & 0.497 & 0.653 & 0.832 & 0.649 & 0.675 & 0.625\\
  & Hist$_{200}$ \cite{sharma_3d_2021}& 0.491 & 0.629 & 0.698 & 0.351 & 0.746 & 0.501 & 0.663 & 0.841 & 0.652 & 0.695 & 0.627\\
  & Euclidean & 0.655 & 0.631 & 0.743 & 0.615 & 0.803 & 0.528 & 0.726 & 0.875 & 0.762 & 0.695 & 0.703\\
  & $\textbf{Ours}_{100}$ & 0.669 & \underline{\textbf{0.642}} & \underline{\textbf{0.808}} & 0.714 & \underline{0.812} & 0.582 & 0.748 & 0.897 & 0.750 & \underline{\textbf{0.733}} & 0.736\\
  & $\textbf{Ours}_{200}$ & \underline{\textbf{0.702}} & 0.630 & 0.728 & \underline{\textbf{0.735}} & \underline{0.812} & \underline{\textbf{0.701}} & \underline{\textbf{0.780}} & \underline{\textbf{0.914}} & \underline{\textbf{0.767}} & 0.713 & \underline{\textbf{0.748}}\\
\bottomrule
\end{tabularx}
    \label{tab:anomaly}
\end{table*}
\begin{table}[t]
\footnotesize
\caption{Accuracy [\%] of human BSE. All the matching methods are applied after RANSAC registration with spectral scaling, when not specified, or CMM + CPD Non-Rigid (\textit{NR}) registration. The overall best performing methods are highlighted in boldface.}
    \centering
\begin{tabularx}{0.45\textwidth}{l|YYYY|l}
\toprule
 Method & Femur & Hip & Fibula & Tibia & Mean $\uparrow$\\
 \midrule  
 Chamfer (\textit{NR}) & 56.58 & 86.58 & 57.89 & 74.74 & 68.95\\
 
 Hausdorff (\textit{NR}) & 57.37 & 87.63 & 59.47 & 73.68 & 69.54\\
 
 Hausdorff & 73.52 & 97.95 & 59.30 & 72.52 & 75.82\\
  
 Chamfer & 71.64 & 98.65 & 64.62 & 74.47 & 77.35\\

 FPFH \cite{rusu_fast_2009} & 68.88 & 96.67 & 66.69 & 78.35 & 77.65\\ %
 
 MAP$_{20}$ \cite{ma_laplacian_2024} & 76.05 & 98.54 & 71.32 & 74.21 & 80.13 \\

 Hist$_{20}$  \cite{sharma_3d_2021} & 77.63 & 98.42 & 69.47 & 77.37 & 80.72 \\
 
 $\textbf{Ours}_{20}$ & 78.79 & \textbf{98.78} & 71.28 & 78.46 & 81.83\\

  $\textbf{Ours}_{10}$ & \textbf{79.68} & 97.76 & \textbf{73.47} & \textbf{78.66} & \textbf{82.39}\\
  
\bottomrule
\end{tabularx}
    \label{tab:benchmark_results}
\end{table}

\textbf{Bone Side Estimation.} The proposed global shape matching score is used to perform automatic BSE. 
Assuming general unknown initial frames, we used the Principal Component Analysis (PCA) to generate a mirrored version of the reference surface, i.e. a synthetic contralateral bone. Namely, we identified that the bilateral symmetry of human and animal bodies is equivalent to a mirroring around the second principal component on the vast majority of bones, such as the ones considered in this study. Hence, given the three point clouds, target, source and mirrored source, the Coupled Laplacian of \cref{fig:coupled-laplacian} can be applied. More specifically, we selected the Random Sample Consensus (RANSAC) \cite{fischler_random_1981}, preceded by a spectral scaling, as rigid registration method to handle varying bone
lengths without altering local structures. The scaling is performed using only the information carried by the Fiedeler vector, i.e. the eigenvector $\pmb{\phi}_1$ corresponding to the first non-zero eigenvalue $\lambda_1$. A rough estimation of the bones length, independently of the Euclidean frame, is given by the distance between the points corresponding to minimum and maximum values of the Fiedler vectors. Thanks to that, one of the two bones can be scaled in order to match the length of the other and improve the RANSAC registration. The three aligned $m$-dimensional embeddings, derived from the eigendecomposition of the Coupled Laplacian, are then used as in \cref{eq:grassman} to retrieve the best matching source and, consequently, the target side. Note that, the proposed BSE method is fully independent of the target and reference frames and therefore can be applied directly on the segmentation obtained from a medical image, without any previous knowledge, in a fully automated pipeline. In alternative, one can using two different bone references, one left and one right, avoiding the mirroring step, or apply the mirroring on the target shape while letting the source unchanged. Full details of the described algorithm are provided in the Supplementary Material.
 
We generated a benchmark to test the BSE task by collecting several bone surface point cloud data from public datasets \cite{fischer_robust_2020, keast_geometric_2023, nolte_non-linear_2016}. Namely, from \textit{Fisher et al.} \cite{fischer_robust_2020} we extracted Femur and Hip structures, from SSM-Tibia \cite{keast_geometric_2023} only right Tibia and Fibula, and from the Imperial College London (ICL) \cite{nolte_non-linear_2016} we collected Femur, Tibia and Fibula point clouds. The detailed composition of the benchmark is reported in \cref{tab:benchmark}. Such a variety of shapes, which are acquired using different methods, makes the task more challenging, including some intra-clinic variability. 
In our experiments, for a given bone class, we performed cross-testing by selecting each shape once as source and the others as targets on which the side has to be inferred. Therefore, the average accuracy of all the experiments, achieved when predicting the correct body side, i.e. binary classification, takes into account the variability of the source shape. The robustness of the method with respect to the reference is discussed in the Supplementary Material.

\begin{figure}
    \centering
    \includegraphics[width=0.39\textwidth]{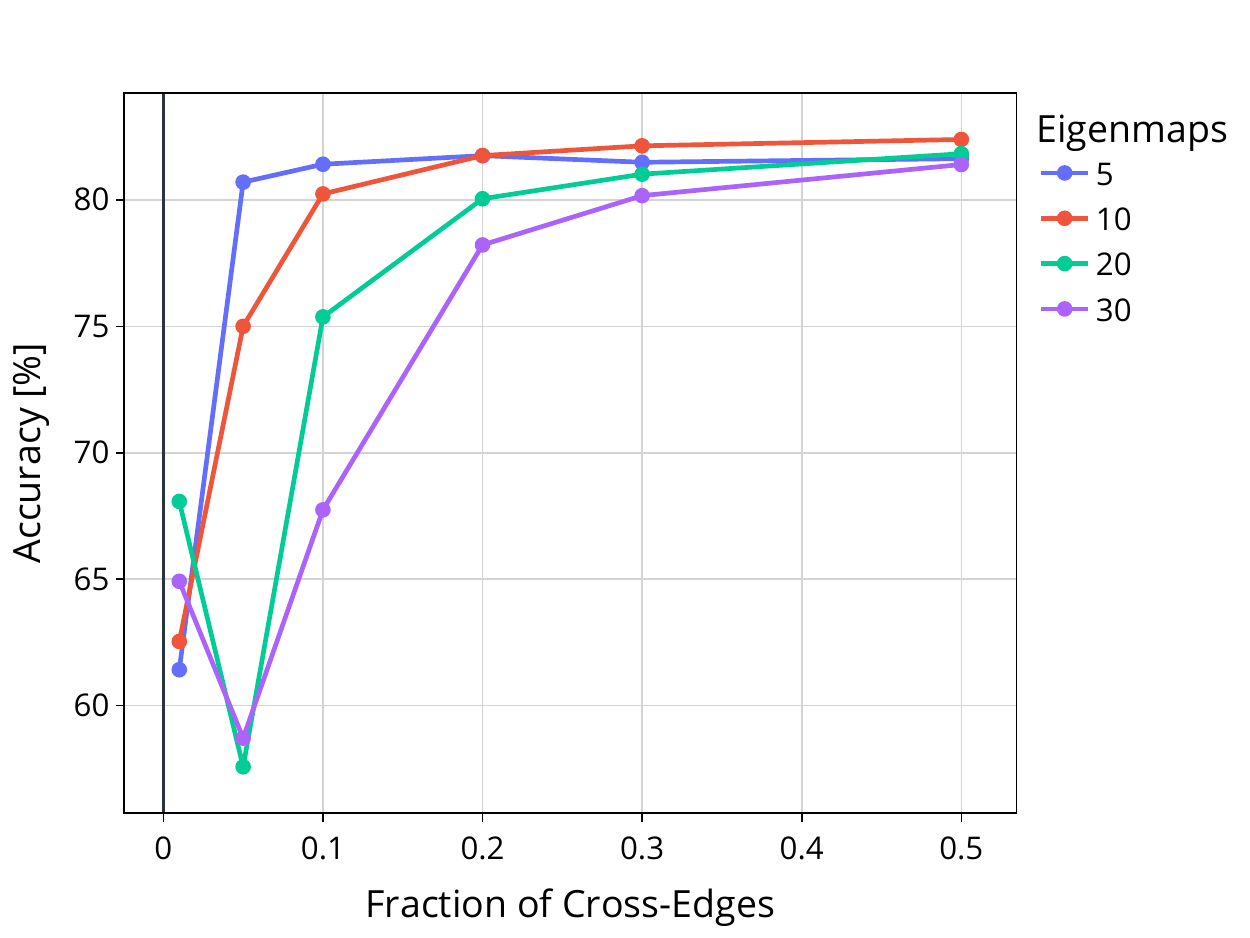}
    \caption{Average BSE accuracy with respect to the fraction of cross-edges used to build the Coupled Laplacian and the number of eigenmaps used for matching.}
    \label{fig:BSE_plot}
\end{figure}
\begin{figure*}
    \centering
    \includegraphics[width=0.615\textwidth]{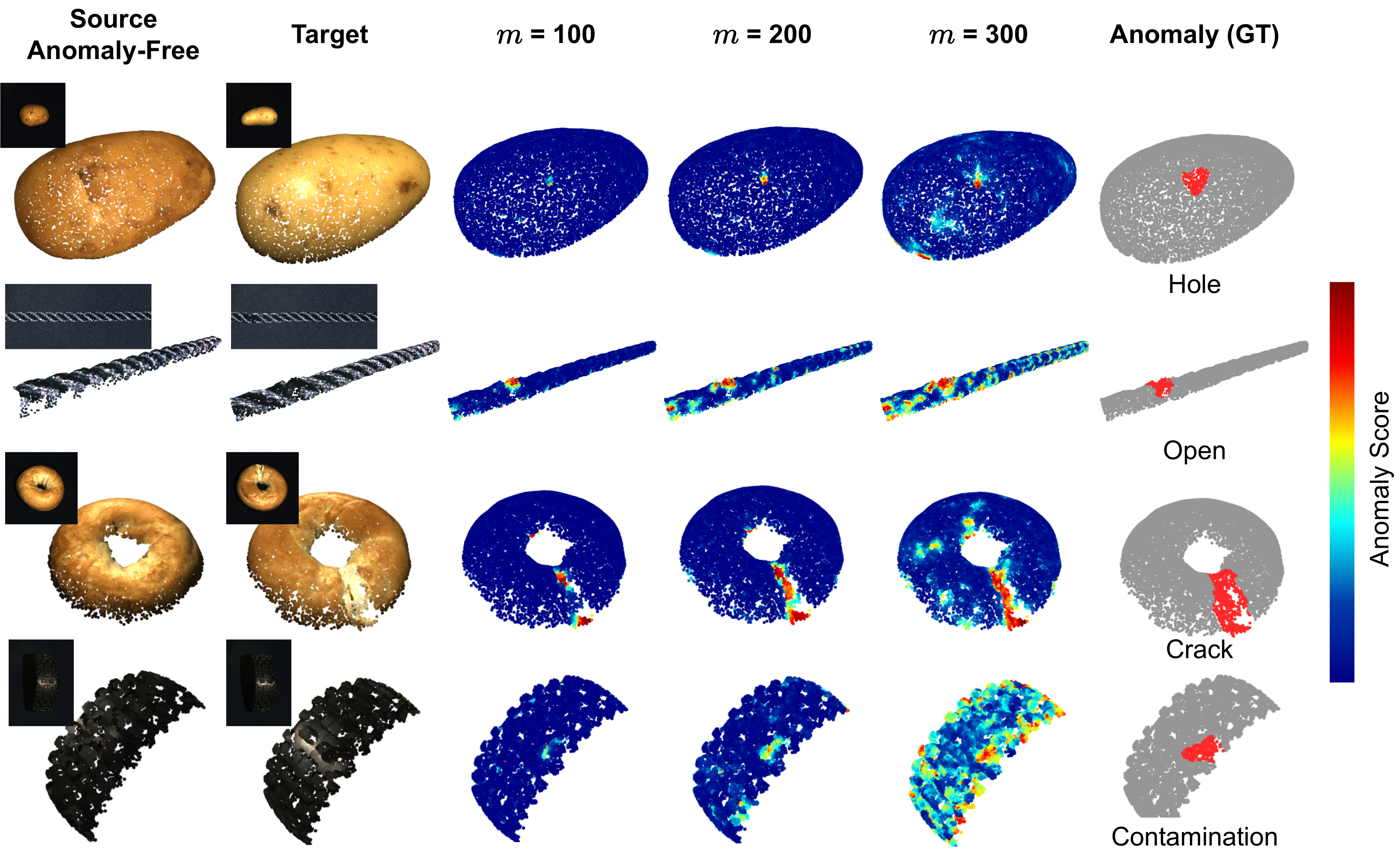}
    \caption{Graphical comparison of 3D anomaly localization using different numbers of aligned eigenmaps, $m$. Tuning the dimension of the embeddings computed with the Coupled Laplacian, it is possible to decide the extent and size of local surface differences that are detected.}
    \label{fig:maps_comparison}
\end{figure*}

\textbf{Anomaly Localization.} We tested the proposed approach for the 3D anomaly detection task in the MVTec 3D-AD dataset \cite{bergmann_mvtec_2022}. This recent dataset was designed for the unsupervised detection of anomalies in point clouds of industrially manufactured products. It contains over 4000 high-resolution 3D scans of 10 object categories. In our experiments, we chose one anomaly-free training sample for each class as source and we used the point-wise distances of the aligned eigenmaps between this shape and each target in the test set as anomaly scores. We performed the pre-registration using affine Coherent Point Drift (CPD) \cite{myronenko_point-set_2010}, which is more suited for objects like the ones included in the MVTec 3D-AD, e.g. length and thickness variations are better captured by an affine registration rather than rigid. In order to speed-up the computation, we pre-processed the point clouds by removing their flat backgrounds with a threshold on the $z$-axis after a 3-dimensional PCA and, without losing generality, we sub-sampled the number of foreground points to a maximum of $13000$. Finally, to compare the obtained anomaly scores with the Ground Truth (GT) image, we projected back the points to the original 2D plane and, if any sub-sampling was performed, we applied a dilation with a structuring element of size the inverse of the sampling factor.

\subsection{Experiment results} 
\label{sec:results}
\textbf{Bone Side Estimation.} In \cref{fig:BSE_plot}, the average accuracy of BSE is depicted based on the fraction of artificial cross-edges added to create the coupled graph and the dimension of the spectral embeddings, $l$ and $m$, respectively. We can observe that the more coupled eigenmaps we aim to utilize for computing a global similarity score, the greater the need for additional cross-connections to ensure the reliability of the results. This occurs because, when graphs are weakly coupled with only a few cross-connections, the Coupled Laplacian captures limited intra-shape characteristics and predominantly emphasizes individual local geometries. In fact, when $l=0$ and therefore the coupled graph is not fully connected, the eigenmaps obtained using the Coupled Laplacian are equivalent to the ones computed independently on each single graph, if the eigenvalues are the same (proof in the Supplmentary Material). On the other hand, not too many cross-edges, nor eigenmaps, are needed to obtain meaningful global matching scores and achieve good performances in the BSE task. In \cref{tab:benchmark_results} a quantitative comparison with other methods of human BSE accuracy is reported. For a fair comparison, all methods are performed after RANSAC registration of the sources. Hausdorff and Chamfer discrepancies are calculated on the Euclidean coordinates of the points, for completeness, also in case of 2-Step non-rigid GMM + CPD registration ($2000$x slower than RANSAC). 
Our method, using $10$ and $20$ eigenmodes with $l=0.5$ fraction of cross-connections, outperforms the other techniques achieving higher accuracy both on the single bones and in average. Hence, the description provided by the aligned eigenmaps, obtained with Coupled Laplacian, is more aware of local details than the other methods. 

\textbf{Anomaly Localization.} \cref{tab:anomaly} lists quantitative results of each evaluated method for the localization of anomalies. For each category, the normalized area under the Per-Region
Overlap (PRO) curve with an upper integration limit of $0.3$ \cite{bergmann_mvtec_2022}, as well as, the mean performance, are reported. 
Performance of Generative Adversarial Network (GAN), Autoencoder (AE) and Variation Model (VM) on the same test set are provided by the dataset authors \cite{bergmann_mvtec_2022}. Furthermore, we tested restricted GPS \cite{rustamov_laplace-beltrami_2007}, eigenmaps histogram matching \cite{mateus_articulated_2008, sharma_3d_2021}, MAP \cite{ma_laplacian_2024} and our method, all with the same pre-processing and source shapes, using $100$ and $200$ eigenmaps. In order to obtain denser and more precise anomaly localization, the coupled graph is built using the whole set of points as cross-connections, i.e. $l=1$. Furthermore, we include the result obtained using as anomaly score for each target point the normalized Euclidean distance of the nearest point in the source geometry, both using rigid and non-rigid registration. Our method, using only 3D information, outperforms all the other techniques. Moreover, we obtained better results than deep-learning methods having RGB and Depth in combination as input. Interestingly, the point-wise similarity computed on eigenmaps not properly aligned (GPS and Hist) is worst than just considering Euclidean distances between points of registered shapes, making worthless the computation of spectral embeddings. In \cref{fig:maps_comparison} we compare the qualitative results obtained  using different numbers of eigenmaps to score the point-wise similarities. Using a smaller $m$, the proposed technique is prone to individuate only small regions with highly dissimilar local geometries. By increasing the size of the spectral embeddings, we can identify larger and more subtle surface differences, even if they are not necessarily classified as anomalies in the GT. For instance, the potato and bagel surfaces have some natural irregularities, with respect to the source, that are not highlighted using $100$ and $200$-dimensional embeddings, but are instead detected with $300$ eigenmaps. This concept is linked to the \textit{modal length}, which we define in the Supplementary Material. The trade-off between number of maps and extent of differences detected is interesting to tune the method to other tasks requiring specific attention to identify surface dissimilarities. 

\begin{figure}
    \centering
    \includegraphics[width=0.45\textwidth]{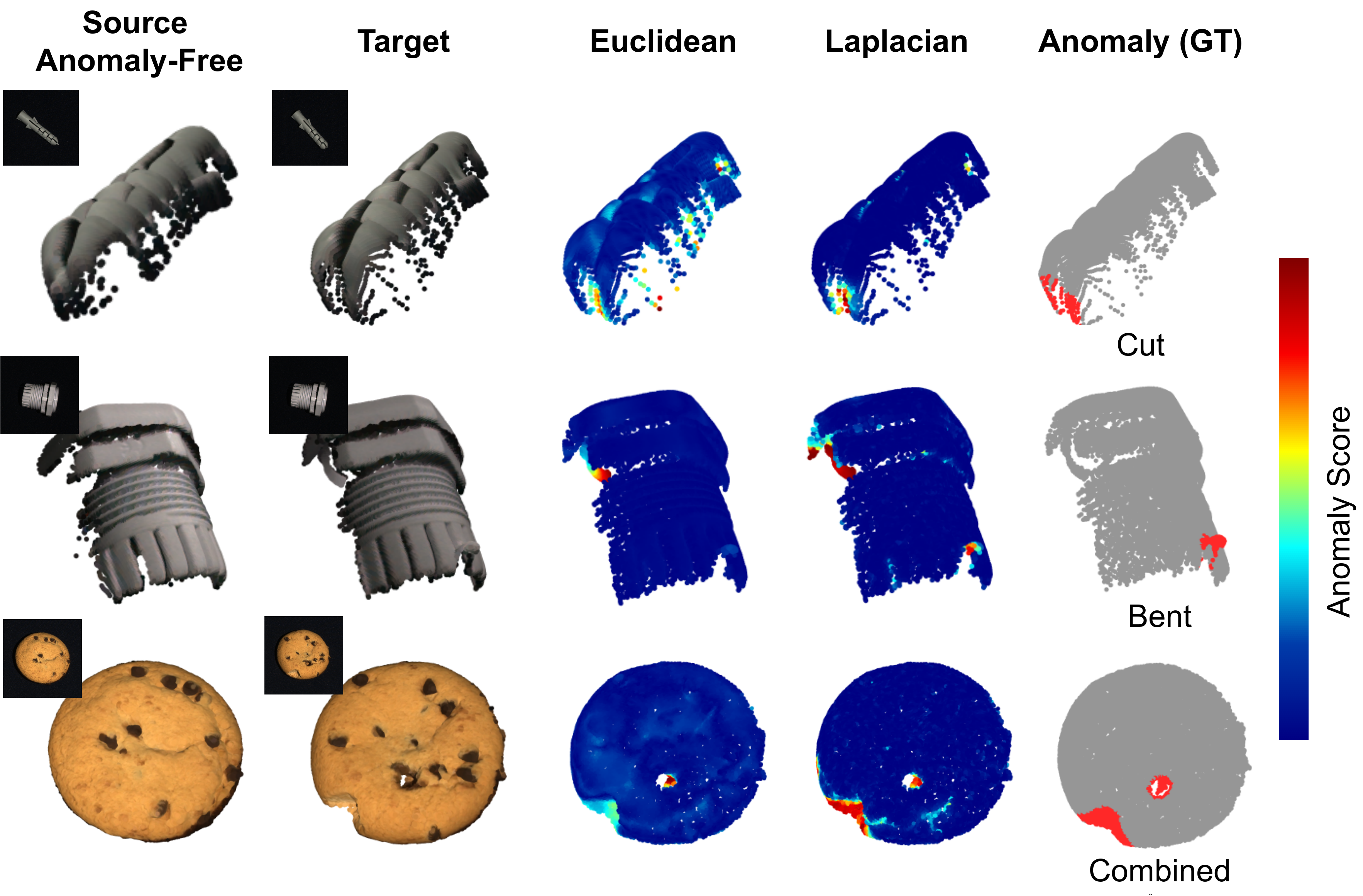}
    \caption{Comparison of the anomaly score obtained using the Coupled Laplacian technique, with $m=200$ and $l=1$, and Euclidean distance, both after affine CPD alignment.}
    \label{fig:euclidean_comparison}
\end{figure}
\textbf{Ablation Study.} 
To further motivate the preference for Coupled Laplacian spectral space over the Euclidean one, in \cref{fig:euclidean_comparison}, we display comparisons of point-wise distances (anomaly scores) between source and target in both spaces, after the same registration. 
In the first and second row the better resistance to noise and outliers in the spectral space is highlighted. The bending of the cable gland, i.e. the anomaly, is barely detected in the Euclidean space because of noisy points on the back. 
Instead, with the aligned eigenmaps comparison, the anomaly, as well as the noise, is correctly localized. This is because both are local differences, with respect to the reference, and, therefore, they can be easily captured with coupled eigenmaps. In addition, in the third row, we show that our technique provides a sharper anomaly split of the cookie point cloud, which simplifies the anomaly detection task. Nevertheless, the usage of the proposed landmarks-free Coupled Laplacian is limited to tasks in which a correspondence between target and source shapes can be defined by mean of a registration process. Problems including graph isomorphism, such as articulated shape matching \cite{mateus_articulated_2008, sharma_3d_2021}, or non-rigid shape matching, can be still solved using our method, without relying on the rigid alignment, if some cross-connection are defined a-priori. In this case, landmarks and their neighbouring points would be most likely the easiest option to connect separate graphs and obtain aligned spectral embeddings for further processing. Moreover, a wrong registration of the shapes, due to randomness and errors in the selected registration algorithm, may affect the alignment of the $m$-dimensional embeddings derived from the Coupled Laplacian. 
We report here some statistics on registration failures in BSE (failure rate [\%] / BSE accuracy on failures [\%]), using different references, for femur 6.84 $ \pm$ 7.82 / 96.67 $\pm$ 11.30 (high symmetry) and hip 0.0 $\pm$ 0.0 (low symmetry). Moreover, to test robustness, we added white Gaussian rotation and translation noise, i.i.d. in the 3 axis post-registration, obtaining (Rot. SD [$^{\circ}$] / Trans. SD [mm] /  Acc. Drop [\%]): 2.5 / 5 / 8.16 - 5 / 5 / 15.13 - 7.5 / 10 / 18.16. Given the low registration failure rates and high robustness to noise, we can rely on the proposed coupling algorithm, making the method independent of the acquisition frames. Moreover, the method can well generalize to any chiral shape matching problem and to tasks like identification of topological noise in graphs and partial matching (see example in Supplementary Material).
\section{Conclusions}

We presented a versatile method to perform 3D rigid point cloud matching, both globally and locally, by using aligned eigenspaces for two or more similar shapes without prior knowledge on the local frames or specif markers on them. The introduction of the Coupled Laplacian operator enables the generation of aligned eigenmaps relying on cross-connections added between graphs of registered geometries. We introduced a novel task consisting in the automatic detection of a bone side, i.e. Bone Side Estimation (BSE), and we proposed a benchmark to test it. Leveraging global similarities of eigenmaps derived from the Coupled Laplacian, we define a full pipeline to perform BSE on arbitrary bone surfaces segmented from a generic medical image. Moreover, we tested the ability of the proposed method in capturing local surface differences by performing 3D anomaly detection on the MVTec 3D-AD dataset. The proposed technique outperforms other methods on the two tasks, therefore capturing better both global and local similarities, thanks to the matching performed on correctly aligned spectral spaces. Beside the applications showcased here, we believe that our work can help several tasks in many fields in which the localization of local difference is crucial for matching, detection or retrieval. Future extensions could explore generalizations of the technique, not only for rigid matching but also to address challenges such as graph isomorphism. This approach opens new avenues for advancing the State-of-The-Art in 3D shape analysis and matching techniques.
\section{Acknowledgements}
This project has received funding from the European Union’s Horizon 2020 research and innovation programme under the Marie Skłodowska-Curie grant agreement No 945304-Cofund AI4theSciences hosted by PSL University.
This work was granted access to the HPC/AI resources of IDRIS under the allocation 2022-AD011013902 made by GENCI.
{
    \small
    \bibliographystyle{ieeenat_fullname}
    \bibliography{references}

\begin{thebibliography}{79}
\providecommand{\natexlab}[1]{#1}
\providecommand{\url}[1]{\texttt{#1}}
\expandafter\ifx\csname urlstyle\endcsname\relax
  \providecommand{\doi}[1]{doi: #1}\else
  \providecommand{\doi}{doi: \begingroup \urlstyle{rm}\Url}\fi

\bibitem[Aubry et~al.(2011)Aubry, Schlickewei, and Cremers]{aubry_wave_2011}
Mathieu Aubry, Ulrich Schlickewei, and Daniel Cremers.
\newblock The wave kernel signature: {A} quantum mechanical approach to shape analysis.
\newblock In \emph{2011 {IEEE} {International} {Conference} on {Computer} {Vision} {Workshops} ({ICCV} {Workshops})}, pages 1626--1633, Barcelona, Spain, 2011. IEEE.

\bibitem[Bandovic et~al.(2023)Bandovic, Holme, Black, and Futterman]{bandovic_anatomy_2023}
Ivan Bandovic, Matthew~R. Holme, Asa~C. Black, and Bennett Futterman.
\newblock Anatomy, {Bone} {Markings}.
\newblock In \emph{{StatPearls}}. StatPearls Publishing, Treasure Island (FL), 2023.

\bibitem[Baum et~al.(2021)Baum, Hu, and Barratt]{baum_real-time_2021}
Zachary M~C Baum, Yipeng Hu, and Dean~C Barratt.
\newblock Real-time multimodal image registration with partial intraoperative point-set data.
\newblock \emph{Medical Image Analysis}, 74:\penalty0 102231, 2021.

\bibitem[Belkin and Niyogi(2003)]{belkin_laplacian_2003}
Mikhail Belkin and Partha Niyogi.
\newblock Laplacian {Eigenmaps} for {Dimensionality} {Reduction} and {Data} {Representation}.
\newblock \emph{Neural Computation}, 15\penalty0 (6):\penalty0 1373--1396, 2003.

\bibitem[Belongie et~al.(2002)Belongie, Malik, and Puzicha]{belongie_shape_2002}
S. Belongie, J. Malik, and J. Puzicha.
\newblock Shape matching and object recognition using shape contexts.
\newblock \emph{IEEE Transactions on Pattern Analysis and Machine Intelligence}, 24\penalty0 (4):\penalty0 509--522, 2002.
\newblock Conference Name: IEEE Transactions on Pattern Analysis and Machine Intelligence.

\bibitem[Bergmann and Sattlegger(2023)]{bergmann_anomaly_2023}
Paul Bergmann and David Sattlegger.
\newblock Anomaly {Detection} in {3D} {Point} {Clouds} using {Deep} {Geometric} {Descriptors}.
\newblock In \emph{2023 {IEEE}/{CVF} {Winter} {Conference} on {Applications} of {Computer} {Vision} ({WACV})}, pages 2612--2622, Waikoloa, HI, USA, 2023. IEEE.

\bibitem[Bergmann et~al.(2022)Bergmann, Jin, Sattlegger, and Steger]{bergmann_mvtec_2022}
Paul Bergmann, Xin Jin, David Sattlegger, and Carsten Steger.
\newblock The {MVTec} {3D}-{AD} {Dataset} for {Unsupervised} {3D} {Anomaly} {Detection} and {Localization}.
\newblock In \emph{Proceedings of the 17th {International} {Joint} {Conference} on {Computer} {Vision}, {Imaging} and {Computer} {Graphics} {Theory} and {Applications}}, pages 202--213, 2022.
\newblock arXiv:2112.09045 [cs].

\bibitem[Besl and McKay(1992)]{besl_method_1992}
P.J. Besl and Neil~D. McKay.
\newblock A method for registration of 3-{D} shapes.
\newblock \emph{IEEE Transactions on Pattern Analysis and Machine Intelligence}, 14\penalty0 (2):\penalty0 239--256, 1992.
\newblock Conference Name: IEEE Transactions on Pattern Analysis and Machine Intelligence.

\bibitem[Bickel et~al.(2023)Bickel, Schleich, and Wartzack]{bickel_novel_2023}
S. Bickel, B. Schleich, and S. Wartzack.
\newblock A {Novel} {Shape} {Retrieval} {Method} for {3D} {Mechanical} {Components} {Based} on {Object} {Projection}, {Pre}-{Trained} {Deep} {Learning} {Models} and {Autoencoder}.
\newblock \emph{Computer-Aided Design}, 154:\penalty0 103417, 2023.

\bibitem[Birdal and Ilic(2015)]{birdal_point_2015}
Tolga Birdal and Slobodan Ilic.
\newblock Point {Pair} {Features} {Based} {Object} {Detection} and {Pose} {Estimation} {Revisited}.
\newblock In \emph{2015 {International} {Conference} on {3D} {Vision}}, pages 527--535, 2015.

\bibitem[Bronstein and Kokkinos(2010)]{bronstein_scale-invariant_2010}
Michael~M. Bronstein and Iasonas Kokkinos.
\newblock Scale-invariant heat kernel signatures for non-rigid shape recognition.
\newblock In \emph{2010 {IEEE} {Computer} {Society} {Conference} on {Computer} {Vision} and {Pattern} {Recognition}}, pages 1704--1711, 2010.
\newblock ISSN: 1063-6919.

\bibitem[Corballis(2020)]{corballis_bilaterally_2020}
Michael~C. Corballis.
\newblock Bilaterally {Symmetrical}: {To} {Be} or {Not} to {Be}?
\newblock \emph{Symmetry}, 12\penalty0 (3):\penalty0 326, 2020.
\newblock Number: 3 Publisher: Multidisciplinary Digital Publishing Institute.

\bibitem[Deng et~al.(2018{\natexlab{a}})Deng, Birdal, and Ilic]{deng_ppf-foldnet_2018}
Haowen Deng, Tolga Birdal, and Slobodan Ilic.
\newblock {PPF}-{FoldNet}: {Unsupervised} {Learning} of {Rotation} {Invariant} {3D} {Local} {Descriptors}, 2018{\natexlab{a}}.
\newblock arXiv:1808.10322 [cs].

\bibitem[Deng et~al.(2018{\natexlab{b}})Deng, Birdal, and Ilic]{deng_ppfnet_2018}
Haowen Deng, Tolga Birdal, and Slobodan Ilic.
\newblock {PPFNet}: {Global} {Context} {Aware} {Local} {Features} for {Robust} {3D} {Point} {Matching}, 2018{\natexlab{b}}.
\newblock arXiv:1802.02669 [cs].

\bibitem[Eckart et~al.(2018)Eckart, Kim, and Kautz]{eckart_fast_2018}
Ben Eckart, Kihwan Kim, and Jan Kautz.
\newblock Fast and {Accurate} {Point} {Cloud} {Registration} using {Trees} of {Gaussian} {Mixtures}, 2018.
\newblock arXiv:1807.02587 [cs].

\bibitem[Elbaz et~al.(2017)Elbaz, Avraham, and Fischer]{elbaz_3d_2017}
Gil Elbaz, Tamar Avraham, and Anath Fischer.
\newblock {3D} {Point} {Cloud} {Registration} for {Localization} {Using} a {Deep} {Neural} {Network} {Auto}-{Encoder}.
\newblock In \emph{2017 {IEEE} {Conference} on {Computer} {Vision} and {Pattern} {Recognition} ({CVPR})}, pages 2472--2481, Honolulu, HI, 2017. IEEE.

\bibitem[Figueroa et~al.(2023)Figueroa, Figueroa, Guiloff, Putnis, Fritsch, and Itriago]{figueroa_navigation_2023}
Francisco Figueroa, David Figueroa, Rodrigo Guiloff, Sven Putnis, Brett Fritsch, and Minerva Itriago.
\newblock Navigation in anterior cruciate ligament reconstruction: {State} of the art.
\newblock \emph{Journal of ISAKOS}, 8\penalty0 (1):\penalty0 47--53, 2023.

\bibitem[Fischer et~al.(2020)Fischer, Grothues, Habor, de~la Fuente, and Radermacher]{fischer_robust_2020}
Maximilian C.~M. Fischer, Sonja A. G.~A. Grothues, Juliana Habor, Matías de~la Fuente, and Klaus Radermacher.
\newblock A robust method for automatic identification of femoral landmarks, axes, planes and bone coordinate systems using surface models.
\newblock \emph{Scientific Reports}, 10\penalty0 (1):\penalty0 20859, 2020.
\newblock Number: 1 Publisher: Nature Publishing Group.

\bibitem[Fischler and Bolles(1981)]{fischler_random_1981}
Martin~A. Fischler and Robert~C. Bolles.
\newblock Random sample consensus: a paradigm for model fitting with applications to image analysis and automated cartography.
\newblock \emph{Communications of the ACM}, 24\penalty0 (6):\penalty0 381--395, 1981.

\bibitem[Gao and Tedrake(2019)]{gao_filterreg_2019}
Wei Gao and Russ Tedrake.
\newblock {FilterReg}: {Robust} and {Efficient} {Probabilistic} {Point}-{Set} {Registration} using {Gaussian} {Filter} and {Twist} {Parameterization}, 2019.
\newblock arXiv:1811.10136 [cs].

\bibitem[Gao et~al.(2023)Gao, Yuan, Ku, Veltkamp, Zamanakos, Tsochatzidis, Amanatiadis, Pratikakis, Panou, Romanelis, Fotis, Arvanitis, and Moustakas]{gao_shrec_2023}
Yang Gao, Honglin Yuan, Tao Ku, Remco~C. Veltkamp, Georgios Zamanakos, Lazaros Tsochatzidis, Angelos Amanatiadis, Ioannis Pratikakis, Aliki Panou, Ioannis Romanelis, Vlassis Fotis, Gerasimos Arvanitis, and Konstantinos Moustakas.
\newblock {SHREC} 2023: {Point} cloud change detection for city scenes.
\newblock \emph{Computers \& Graphics}, 115:\penalty0 35--42, 2023.

\bibitem[Ge and Fan(2015)]{ge_articulated_2015}
Song Ge and Guoliang Fan.
\newblock Articulated {Non}-{Rigid} {Point} {Set} {Registration} for {Human} {Pose} {Estimation} from {3D} {Sensors}.
\newblock \emph{Sensors}, 15\penalty0 (7):\penalty0 15218--15245, 2015.
\newblock Number: 7 Publisher: Multidisciplinary Digital Publishing Institute.

\bibitem[Ghojogh et~al.(2022)Ghojogh, Ghodsi, Karray, and Crowley]{ghojogh_laplacian-based_2022}
Benyamin Ghojogh, Ali Ghodsi, Fakhri Karray, and Mark Crowley.
\newblock Laplacian-{Based} {Dimensionality} {Reduction} {Including} {Spectral} {Clustering}, {Laplacian} {Eigenmap}, {Locality} {Preserving} {Projection}, {Graph} {Embedding}, and {Diffusion} {Map}: {Tutorial} and {Survey}, 2022.
\newblock arXiv:2106.02154 [cs, stat].

\bibitem[Gojcic et~al.(2019)Gojcic, Zhou, Wegner, and Wieser]{gojcic_perfect_2019}
Zan Gojcic, Caifa Zhou, Jan~D. Wegner, and Andreas Wieser.
\newblock The {Perfect} {Match}: {3D} {Point} {Cloud} {Matching} {With} {Smoothed} {Densities}.
\newblock In \emph{2019 {IEEE}/{CVF} {Conference} on {Computer} {Vision} and {Pattern} {Recognition} ({CVPR})}, pages 5540--5549, Long Beach, CA, USA, 2019. IEEE.

\bibitem[Guo et~al.(2013)Guo, Sohel, Bennamoun, Lu, and Wan]{guo_rotational_2013}
Yulan Guo, Ferdous Sohel, Mohammed Bennamoun, Min Lu, and Jianwei Wan.
\newblock Rotational {Projection} {Statistics} for {3D} {Local} {Surface} {Description} and {Object} {Recognition}.
\newblock \emph{International Journal of Computer Vision}, 105\penalty0 (1):\penalty0 63--86, 2013.
\newblock arXiv:1304.3192 [cs].

\bibitem[Guo et~al.(2016)Guo, Bennamoun, Sohel, Lu, Wan, and Kwok]{guo_comprehensive_2016}
Yulan Guo, Mohammed Bennamoun, Ferdous Sohel, Min Lu, Jianwei Wan, and Ngai~Ming Kwok.
\newblock A {Comprehensive} {Performance} {Evaluation} of {3D} {Local} {Feature} {Descriptors}.
\newblock \emph{International Journal of Computer Vision}, 116\penalty0 (1):\penalty0 66--89, 2016.

\bibitem[Holló and Novák(2012)]{hollo_manoeuvrability_2012}
Gábor Holló and Mihály Novák.
\newblock The manoeuvrability hypothesis to explain the maintenance of bilateral symmetry in animal evolution.
\newblock \emph{Biology Direct}, 7\penalty0 (1):\penalty0 22, 2012.

\bibitem[Hu and Hua(2009)]{hu_salient_2009}
Jiaxi Hu and Jing Hua.
\newblock Salient spectral geometric features for shape matching and retrieval.
\newblock \emph{The Visual Computer}, 25\penalty0 (5):\penalty0 667--675, 2009.

\bibitem[Huang et~al.(2017)Huang, Kalogerakis, Chaudhuri, Ceylan, Kim, and Yumer]{huang_learning_2017}
Haibin Huang, Evangelos Kalogerakis, Siddhartha Chaudhuri, Duygu Ceylan, Vladimir~G. Kim, and Ersin Yumer.
\newblock Learning {Local} {Shape} {Descriptors} from {Part} {Correspondences} {With} {Multi}-view {Convolutional} {Networks}, 2017.
\newblock arXiv:1706.04496 [cs].

\bibitem[Jian and Vemuri(2011)]{jian_robust_2011}
Bing Jian and Baba~C. Vemuri.
\newblock Robust {Point} {Set} {Registration} {Using} {Gaussian} {Mixture} {Models}.
\newblock \emph{IEEE Transactions on Pattern Analysis and Machine Intelligence}, 33\penalty0 (8):\penalty0 1633--1645, 2011.
\newblock Conference Name: IEEE Transactions on Pattern Analysis and Machine Intelligence.

\bibitem[Keast et~al.(2023)Keast, Bonacci, and Fox]{keast_geometric_2023}
Meghan Keast, Jason Bonacci, and Aaron Fox.
\newblock Geometric variation of the human tibia-fibula: a public dataset of tibia-fibula surface meshes and statistical shape model.
\newblock \emph{PeerJ}, 11:\penalty0 e14708, 2023.
\newblock Publisher: PeerJ Inc.

\bibitem[Kobayashi et~al.(2023)Kobayashi, Gu, Hataya, Mizuno, Miyake, Watanabe, Takahashi, Takamizawa, Yoshida, Nakamura, Kouno, Bolatkan, Kurose, Harada, and Hamamoto]{kobayashi_sketch-based_2023}
Kazuma Kobayashi, Lin Gu, Ryuichiro Hataya, Takaaki Mizuno, Mototaka Miyake, Hirokazu Watanabe, Masamichi Takahashi, Yasuyuki Takamizawa, Yukihiro Yoshida, Satoshi Nakamura, Nobuji Kouno, Amina Bolatkan, Yusuke Kurose, Tatsuya Harada, and Ryuji Hamamoto.
\newblock Sketch-based {Medical} {Image} {Retrieval}, 2023.
\newblock arXiv:2303.03633 [cs].

\bibitem[Lambrechts et~al.(2022)Lambrechts, Wirix-Speetjens, Maes, and Van~Huffel]{lambrechts_artificial_2022}
Adriaan Lambrechts, Roel Wirix-Speetjens, Frederik Maes, and Sabine Van~Huffel.
\newblock Artificial {Intelligence} {Based} {Patient}-{Specific} {Preoperative} {Planning} {Algorithm} for {Total} {Knee} {Arthroplasty}.
\newblock \emph{Frontiers in robotics and AI}, 9:\penalty0 840282, 2022.

\bibitem[Li et~al.(2018)Li, Yang, Wang, and Wang]{li_rigid_2018}
Liang Li, Ming Yang, Chunxiang Wang, and Bing Wang.
\newblock Rigid {Point} {Set} {Registration} {Based} on {Cubature} {Kalman} {Filter} and {Its} {Application} in {Intelligent} {Vehicles}.
\newblock \emph{IEEE Transactions on Intelligent Transportation Systems}, 19\penalty0 (6):\penalty0 1754--1765, 2018.
\newblock Conference Name: IEEE Transactions on Intelligent Transportation Systems.

\bibitem[Lian et~al.(2013)Lian, Godil, Bustos, Daoudi, Hermans, Kawamura, Kurita, Lavoué, Van~Nguyen, Ohbuchi, Ohkita, Ohishi, Porikli, Reuter, Sipiran, Smeets, Suetens, Tabia, and Vandermeulen]{lian_comparison_2013}
Zhouhui Lian, Afzal Godil, Benjamin Bustos, Mohamed Daoudi, Jeroen Hermans, Shun Kawamura, Yukinori Kurita, Guillaume Lavoué, Hien Van~Nguyen, Ryutarou Ohbuchi, Yuki Ohkita, Yuya Ohishi, Fatih Porikli, Martin Reuter, Ivan Sipiran, Dirk Smeets, Paul Suetens, Hedi Tabia, and Dirk Vandermeulen.
\newblock A comparison of methods for non-rigid {3D} shape retrieval.
\newblock \emph{Pattern Recognition}, 46\penalty0 (1):\penalty0 449--461, 2013.

\bibitem[Lian et~al.(2015)Lian, Zhang, Choi, ElNaghy, El-Sana, Furuya, Giachetti, Guler, Lai, Li, Li, Limberger, Martin, Nakanishi, Neto, Nonato, Ohbuchi, Pevzner, Pickup, Rosin, Sharf, Sun, Sun, Tari, Unal, and Wilson]{lian_non-rigid_2015}
Z. Lian, J. Zhang, S. Choi, H. ElNaghy, J. El-Sana, T. Furuya, A. Giachetti, R.~A. Guler, L. Lai, C. Li, H. Li, F.~A. Limberger, R. Martin, R.~U. Nakanishi, A.~P. Neto, L.~G. Nonato, R. Ohbuchi, K. Pevzner, D. Pickup, P. Rosin, A. Sharf, L. Sun, X. Sun, S. Tari, G. Unal, and R.~C. Wilson.
\newblock Non-rigid {3D} shape retrieval.
\newblock In \emph{Proceedings of the 2015 {Eurographics} {Workshop} on {3D} {Object} {Retrieval}}, pages 107--120, Goslar, DEU, 2015. Eurographics Association.

\bibitem[Lotan et~al.(2020)Lotan, Tschider, Sodickson, Caplan, Bruno, Zhang, and Lui]{lotan_medical_2020}
Eyal Lotan, Charlotte Tschider, Daniel~K. Sodickson, Arthur~L. Caplan, Mary Bruno, Ben Zhang, and Yvonne~W. Lui.
\newblock Medical {Imaging} and {Privacy} in the {Era} of {Artificial} {Intelligence}: {Myth}, {Fallacy}, and the {Future}.
\newblock \emph{Journal of the American College of Radiology : JACR}, 17\penalty0 (9):\penalty0 1159--1162, 2020.

\bibitem[Lowe(2004)]{lowe_distinctive_2004}
David~G. Lowe.
\newblock Distinctive {Image} {Features} from {Scale}-{Invariant} {Keypoints}.
\newblock \emph{International Journal of Computer Vision}, 60\penalty0 (2):\penalty0 91--110, 2004.

\bibitem[Ma et~al.(2024)Ma, Wang, and Wang]{ma_laplacian_2024}
Jiangyan Ma, Yifei Wang, and Yisen Wang.
\newblock Laplacian {Canonization}: {A} {Minimalist} {Approach} to {Sign} and {Basis} {Invariant} {Spectral} {Embedding}, 2024.
\newblock arXiv:2310.18716 [cs].

\bibitem[Maeztu~Redin et~al.(2022)Maeztu~Redin, Caroux, Rohan, Pillet, Cermolacce, Trnka, Manassero, Viateau, and Corté]{maeztu_redin_wear_2022}
Deyo Maeztu~Redin, Julien Caroux, Pierre-Yves Rohan, Hélène Pillet, Alexia Cermolacce, Julien Trnka, Mathieu Manassero, Véronique Viateau, and Laurent Corté.
\newblock A wear model to predict damage of reconstructed {ACL}.
\newblock \emph{Journal of the Mechanical Behavior of Biomedical Materials}, page 105426, 2022.

\bibitem[Maiseli et~al.(2017)Maiseli, Gu, and Gao]{maiseli_recent_2017}
Baraka Maiseli, Yanfeng Gu, and Huijun Gao.
\newblock Recent developments and trends in point set registration methods.
\newblock \emph{Journal of Visual Communication and Image Representation}, 46:\penalty0 95--106, 2017.

\bibitem[Mateus et~al.(2008)Mateus, Horaud, Knossow, Cuzzolin, and Boyer]{mateus_articulated_2008}
Diana Mateus, Radu Horaud, David Knossow, Fabio Cuzzolin, and Edmond Boyer.
\newblock Articulated {Shape} {Matching} {Using} {Laplacian} {Eigenfunctions} and {Unsupervised} {Point} {Registration}.
\newblock In \emph{2008 {IEEE} {Conference} on {Computer} {Vision} and {Pattern} {Recognition}}, pages 1--8, 2008.
\newblock arXiv:2012.07340 [cs].

\bibitem[Merris(1994)]{merris_laplacian_1994}
Russell Merris.
\newblock Laplacian matrices of graphs: a survey.
\newblock \emph{Linear Algebra and its Applications}, 197-198:\penalty0 143--176, 1994.

\bibitem[Morita et~al.(2015)Morita, Kobashi, Kashiwa, Nakayama, Kambara, Morimoto, Yoshiya, and Aikawa]{morita_computer-aided_2015}
Kento Morita, Syoji Kobashi, Kaori Kashiwa, Hiroshi Nakayama, Shunichiro Kambara, Masakazu Morimoto, Shinichi Yoshiya, and Satoru Aikawa.
\newblock Computer-aided {Surgical} {Planning} of {Anterior} {Cruciate} {Ligament} {Reconstruction} in {MR} {Images}.
\newblock \emph{Procedia Computer Science}, 60:\penalty0 1659--1667, 2015.

\bibitem[Myronenko and Song(2010)]{myronenko_point-set_2010}
Andriy Myronenko and Xubo Song.
\newblock Point-{Set} {Registration}: {Coherent} {Point} {Drift}.
\newblock \emph{IEEE Transactions on Pattern Analysis and Machine Intelligence}, 32\penalty0 (12):\penalty0 2262--2275, 2010.
\newblock arXiv:0905.2635 [cs].

\bibitem[Nguyen et~al.(2020)Nguyen, Gopalan, Patel, Corsaro, Pavlick, and Tellex]{nguyen_robot_2020}
Thao Nguyen, Nakul Gopalan, Roma Patel, Matt Corsaro, Ellie Pavlick, and Stefanie Tellex.
\newblock Robot {Object} {Retrieval} with {Contextual} {Natural} {Language} {Queries}, 2020.
\newblock arXiv:2006.13253 [cs].

\bibitem[Nolte et~al.(2016)Nolte, Tsang, Zhang, Ding, Kedgley, and Bull]{nolte_non-linear_2016}
Daniel Nolte, Chui~Kit Tsang, Kai~Yu Zhang, Ziyun Ding, Angela~E. Kedgley, and Anthony M.~J. Bull.
\newblock Non-linear scaling of a musculoskeletal model of the lower limb using statistical shape models.
\newblock \emph{Journal of Biomechanics}, 49\penalty0 (14):\penalty0 3576--3581, 2016.

\bibitem[Pilevar(2011)]{pilevar_cbmir_2011}
Abdol~Hamid Pilevar.
\newblock {CBMIR}: {Content}-based {Image} {Retrieval} {Algorithm} for {Medical} {Image} {Databases}.
\newblock \emph{Journal of Medical Signals and Sensors}, 1\penalty0 (1):\penalty0 12--18, 2011.

\bibitem[Qi et~al.(2017{\natexlab{a}})Qi, Su, Mo, and Guibas]{qi_pointnet_2017}
Charles~R. Qi, Hao Su, Kaichun Mo, and Leonidas~J. Guibas.
\newblock {PointNet}: {Deep} {Learning} on {Point} {Sets} for {3D} {Classification} and {Segmentation}, 2017{\natexlab{a}}.
\newblock arXiv:1612.00593 [cs].

\bibitem[Qi et~al.(2017{\natexlab{b}})Qi, Yi, Su, and Guibas]{qi_pointnet2_2017}
Charles~R. Qi, Li Yi, Hao Su, and Leonidas~J. Guibas.
\newblock {PointNet}++: {Deep} {Hierarchical} {Feature} {Learning} on {Point} {Sets} in a {Metric} {Space}, 2017{\natexlab{b}}.
\newblock arXiv:1706.02413 [cs].

\bibitem[Reuter et~al.(2005)Reuter, Wolter, and Peinecke]{reuter_laplace-spectra_2005}
Martin Reuter, Franz-Erich Wolter, and Niklas Peinecke.
\newblock Laplace-spectra as fingerprints for shape matching.
\newblock In \emph{Proceedings of the 2005 {ACM} symposium on {Solid} and physical modeling}, pages 101--106, New York, NY, USA, 2005. Association for Computing Machinery.

\bibitem[Reuter et~al.(2006)Reuter, Wolter, and Peinecke]{reuter_laplacebeltrami_2006}
Martin Reuter, Franz-Erich Wolter, and Niklas Peinecke.
\newblock Laplace–{Beltrami} spectra as ‘{Shape}-{DNA}’ of surfaces and solids.
\newblock \emph{Computer-Aided Design}, 38\penalty0 (4):\penalty0 342--366, 2006.

\bibitem[Reuter et~al.(2009)Reuter, Biasotti, Giorgi, Patanè, and Spagnuolo]{reuter_discrete_2009}
Martin Reuter, Silvia Biasotti, Daniela Giorgi, Giuseppe Patanè, and Michela Spagnuolo.
\newblock Discrete {Laplace}–{Beltrami} operators for shape analysis and segmentation.
\newblock \emph{Computers \& Graphics}, 33\penalty0 (3):\penalty0 381--390, 2009.

\bibitem[Rostami et~al.(2019)Rostami, Bashiri, Rostami, and Yu]{rostami_survey_2019}
R. Rostami, F.~S. Bashiri, B. Rostami, and Z. Yu.
\newblock A {Survey} on {Data}-{Driven} {3D} {Shape} {Descriptors}.
\newblock \emph{Computer Graphics Forum}, 38\penalty0 (1):\penalty0 356--393, 2019.
\newblock \_eprint: https://onlinelibrary.wiley.com/doi/pdf/10.1111/cgf.13536.

\bibitem[Rustamov(2007)]{rustamov_laplace-beltrami_2007}
Raif~M. Rustamov.
\newblock \emph{Laplace-{Beltrami} {Eigenfunctions} for {Deformation} {Invariant} {Shape} {Representation}}.
\newblock The Eurographics Association, 2007.
\newblock Accepted: 2014-01-29T09:43:15Z ISSN: 1727-8384.

\bibitem[Rusu et~al.(2008)Rusu, Blodow, Marton, and Beetz]{rusu_aligning_2008}
Radu~Bogdan Rusu, Nico Blodow, Zoltan~Csaba Marton, and Michael Beetz.
\newblock Aligning point cloud views using persistent feature histograms.
\newblock In \emph{2008 {IEEE}/{RSJ} {International} {Conference} on {Intelligent} {Robots} and {Systems}}, pages 3384--3391, 2008.
\newblock ISSN: 2153-0866.

\bibitem[Rusu et~al.(2009)Rusu, Blodow, and Beetz]{rusu_fast_2009}
Radu~Bogdan Rusu, Nico Blodow, and Michael Beetz.
\newblock Fast {Point} {Feature} {Histograms} ({FPFH}) for {3D} registration.
\newblock In \emph{2009 {IEEE} {International} {Conference} on {Robotics} and {Automation}}, pages 3212--3217, Kobe, 2009. IEEE.

\bibitem[Salti et~al.(2014)Salti, Tombari, and Di~Stefano]{salti_shot_2014}
Samuele Salti, Federico Tombari, and Luigi Di~Stefano.
\newblock {SHOT}: {Unique} signatures of histograms for surface and texture description.
\newblock \emph{Computer Vision and Image Understanding}, 125:\penalty0 251--264, 2014.

\bibitem[Shajahan et~al.(2021)Shajahan, T, and Muthuganapathy]{shajahan_point_2021}
Dimple~A. Shajahan, Mukund~Varma T, and Ramanathan Muthuganapathy.
\newblock Point {Transformer} for {Shape} {Classification} and {Retrieval} of {3D} and {ALS} {Roof} {PointClouds}, 2021.
\newblock arXiv:2011.03921 [cs].

\bibitem[Sharma et~al.(2021)Sharma, Horaud, and Mateus]{sharma_3d_2021}
Avinash Sharma, Radu Horaud, and Diana Mateus.
\newblock {3D} {Shape} {Registration} {Using} {Spectral} {Graph} {Embedding} and {Probabilistic} {Matching}, 2021.
\newblock arXiv:2106.11166 [cs, stat].

\bibitem[Sinko et~al.(2018)Sinko, Kamencay, Hudec, and Benco]{sinko_3d_2018}
Martin Sinko, Patrik Kamencay, Robert Hudec, and Miroslav Benco.
\newblock {3D} registration of the point cloud data using {ICP} algorithm in medical image analysis.
\newblock In \emph{2018 {ELEKTRO}}, pages 1--6, 2018.

\bibitem[Sun et~al.(2009)Sun, Ovsjanikov, and Guibas]{sun_concise_2009}
Jian Sun, Maks Ovsjanikov, and Leonidas Guibas.
\newblock A {Concise} and {Provably} {Informative} {Multi}‐{Scale} {Signature} {Based} on {Heat} {Diffusion}.
\newblock \emph{Computer Graphics Forum}, 28\penalty0 (5):\penalty0 1383--1392, 2009.

\bibitem[Tang and Godil(2012)]{tang_evaluation_2012}
Sarah Tang and Afzal Godil.
\newblock An evaluation of local shape descriptors for {3D} shape retrieval.
\newblock In \emph{Three-{Dimensional} {Image} {Processing} ({3DIP}) and {Applications} {II}}, pages 217--231. SPIE, 2012.

\bibitem[Tang and Tomizuka(2022)]{tang_track_2022}
Te Tang and Masayoshi Tomizuka.
\newblock Track deformable objects from point clouds with structure preserved registration.
\newblock \emph{The International Journal of Robotics Research}, 41\penalty0 (6):\penalty0 599--614, 2022.
\newblock Publisher: SAGE Publications Ltd STM.

\bibitem[Tangelder and Veltkamp(2008)]{tangelder_survey_2008}
Johan W.~H. Tangelder and Remco~C. Veltkamp.
\newblock A survey of content based {3D} shape retrieval methods.
\newblock \emph{Multimedia Tools and Applications}, 39\penalty0 (3):\penalty0 441--471, 2008.

\bibitem[Tombari et~al.(2010)Tombari, Salti, and Di~Stefano]{tombari_unique_2010}
Federico Tombari, Samuele Salti, and Luigi Di~Stefano.
\newblock Unique shape context for 3d data description.
\newblock In \emph{Proceedings of the {ACM} workshop on {3D} object retrieval}, pages 57--62, New York, NY, USA, 2010. Association for Computing Machinery.

\bibitem[Toxvaerd(2021)]{toxvaerd_emergence_2021}
Søren Toxvaerd.
\newblock The {Emergence} of the {Bilateral} {Symmetry} in {Animals}: {A} {Review} and a {New} {Hypothesis}.
\newblock \emph{Symmetry}, 13\penalty0 (2):\penalty0 261, 2021.
\newblock Number: 2 Publisher: Multidisciplinary Digital Publishing Institute.

\bibitem[Wang and Solomon(2019)]{wang_deep_2019}
Yue Wang and Justin~M. Solomon.
\newblock Deep {Closest} {Point}: {Learning} {Representations} for {Point} {Cloud} {Registration}, 2019.
\newblock arXiv:1905.03304 [cs] version: 1.

\bibitem[Wang et~al.(2019)Wang, Guo, Yan, Wang, and Zhang]{wang_robust_2019}
Yiqun Wang, Jianwei Guo, Dong-Ming Yan, Kai Wang, and Xiaopeng Zhang.
\newblock A {Robust} {Local} {Spectral} {Descriptor} for {Matching} {Non}-{Rigid} {Shapes} {With} {Incompatible} {Shape} {Structures}.
\newblock In \emph{2019 {IEEE}/{CVF} {Conference} on {Computer} {Vision} and {Pattern} {Recognition} ({CVPR})}, pages 6224--6233, Long Beach, CA, USA, 2019. IEEE.

\bibitem[Weinmann et~al.(2014)Weinmann, Jutzi, and Mallet]{weinmann_semantic_2014}
Martin Weinmann, Boris Jutzi, and Clément Mallet.
\newblock Semantic {3D} scene interpretation: {A} framework combining optimal neighborhood size selection with relevant features.
\newblock \emph{ISPRS Annals of the Photogrammetry, Remote Sensing and Spatial Information Sciences}, II-3:\penalty0 181--188, 2014.

\bibitem[Wu et~al.(2023)Wu, Fang, Yu, and Yang]{wu_learning_2023}
Hao Wu, Lincong Fang, Qian Yu, and Chengzhuan Yang.
\newblock Learning {Robust} {Point} {Representation} for {3D} {Non}-{Rigid} {Shape} {Retrieval}.
\newblock \emph{IEEE Transactions on Multimedia}, pages 1--15, 2023.
\newblock Conference Name: IEEE Transactions on Multimedia.

\bibitem[Xie et~al.(2017)Xie, Dai, Zhu, Wong, and Fang]{xie_deepshape_2017}
Jin Xie, Guoxian Dai, Fan Zhu, Edward~K. Wong, and Yi Fang.
\newblock {DeepShape}: {Deep}-{Learned} {Shape} {Descriptor} for {3D} {Shape} {Retrieval}.
\newblock \emph{IEEE Transactions on Pattern Analysis and Machine Intelligence}, 39\penalty0 (7):\penalty0 1335--1345, 2017.
\newblock Conference Name: IEEE Transactions on Pattern Analysis and Machine Intelligence.

\bibitem[Xu and Fang(2016)]{xu_shape_2016}
Guoqing Xu and Weiwei Fang.
\newblock Shape retrieval using deep autoencoder learning representation.
\newblock In \emph{2016 13th {International} {Computer} {Conference} on {Wavelet} {Active} {Media} {Technology} and {Information} {Processing} ({ICCWAMTIP})}, pages 227--230, 2016.

\bibitem[Ye and Lim(2016)]{ye_schubert_2016}
Ke Ye and Lek-Heng Lim.
\newblock Schubert {Varieties} and {Distances} between {Subspaces} of {Different} {Dimensions}.
\newblock \emph{SIAM Journal on Matrix Analysis and Applications}, 37\penalty0 (3):\penalty0 1176--1197, 2016.

\bibitem[Yew and Lee(2018)]{yew_3dfeat-net_2018}
Zi~Jian Yew and Gim~Hee Lee.
\newblock {3DFeat}-{Net}: {Weakly} {Supervised} {Local} {3D} {Features} for {Point} {Cloud} {Registration}.
\newblock In \emph{Computer {Vision} – {ECCV} 2018}, pages 630--646, Cham, 2018. Springer International Publishing.

\bibitem[Yi et~al.(2016)Yi, Trulls, Lepetit, and Fua]{yi_lift_2016}
Kwang~Moo Yi, Eduard Trulls, Vincent Lepetit, and Pascal Fua.
\newblock {LIFT}: {Learned} {Invariant} {Feature} {Transform}, 2016.
\newblock arXiv:1603.09114 [cs].

\bibitem[Zeng et~al.(2017)Zeng, Song, Nießner, Fisher, Xiao, and Funkhouser]{zeng_3dmatch_2017}
Andy Zeng, Shuran Song, Matthias Nießner, Matthew Fisher, Jianxiong Xiao, and Thomas Funkhouser.
\newblock {3DMatch}: {Learning} {Local} {Geometric} {Descriptors} from {RGB}-{D} {Reconstructions}, 2017.
\newblock arXiv:1603.08182 [cs].

\bibitem[Zhang and Chen(2001)]{zhang_efficient_2001}
Cha Zhang and Tsuhan Chen.
\newblock Efficient feature extraction for {2D}/{3D} objects in mesh representation.
\newblock In \emph{Proceedings 2001 {International} {Conference} on {Image} {Processing} ({Cat}. {No}.{01CH37205})}, pages 935--938 vol.3, 2001.

\bibitem[Zhang et~al.(2010)Zhang, Van~Kaick, and Dyer]{zhang_spectral_2010}
H. Zhang, O. Van~Kaick, and R. Dyer.
\newblock Spectral {Mesh} {Processing}.
\newblock \emph{Computer Graphics Forum}, 29\penalty0 (6):\penalty0 1865--1894, 2010.

\end{thebibliography}
}

\clearpage
\setcounter{page}{1}
\maketitlesupplementary

\section{Coupled Laplacian}
\label{sec:theory}
We present here some theoretical aspects of the Coupled Laplacian, introduced in \cref{sec:method}, which we proposed to generate aligned spectral embeddings for multiple registered point clouds.

\textbf{Zero Cross-Connections.} The coupled Laplacian matrix has the following structure:
\begin{equation}
\label{eq:coupled_laplacian}
\pmb{L}^C = \pmb{L}^U + \pmb{L}^+
\end{equation}
where $\pmb{L}^U$ is the Laplacian matrix of the global graph without any cross-connection between $\mathcal{G}^\mathcal{T}$ and $\mathcal{G}^{S_k}$, for $k=1, \cdots, N$. Hence, $\pmb{L}^+$ is the Laplacian matrix of the global graph restricted to its cross-connections between sub-graphs. Moreover, $\pmb{L}^+$ enforces the matching constraint in the global eigenproblem.

By construction of the matrix $\pmb{B}^C$ (see \cref{sec:method}), eigenvectors related to $(\mathcal{G}^{S_k})_{k=1,\cdots N}$ and $\mathcal{G}^\mathcal{T}$, that have the same eigenvalue $\lambda$, can be stacked into a global eigenvector $\pmb{\phi}^U$ such that the following property holds
\begin{equation}
\label{eq:uncoupled}
\pmb{L}^U \: \pmb{\phi}^U = \lambda \: \pmb{B}^C \: \pmb{\phi}^U 
\end{equation}
where the above system is block diagonal. Therefore, solving the Coupled Laplacian eigenproblem without cross-connections, i.e. $l=0$ and then $\pmb{L}^+ = \pmb{0}$, it is equivalent of solving separate eigenprobems, when the eigenvalues are the same, for each single component of the global graph. Otherwise, if eigenvalues are not the same, the solution of the coupled eigenproblems with $\pmb{L}^+ = \pmb{0}$ will likely give separate eigenvectors for each sub-graph, that is, vectors with non-zero values only for indices corresponding to a single component. 

\textbf{Ideal Matching.} In the ideal case of perfect matching, when source graphs bare copies of the target graph, whether or not vertices are randomly reordered, the set of $n^\mathcal{T}$ coupled eigenvectors contains $N+1$ copies of target's eigenvectors. For the sake of simplicity, in the following proof, the vertices in the copies of the target graph are not reordered. Therefore, $\mathcal{G}^{S_k} = \mathcal{G}^\mathcal{T}$ and $F^{S_k} = F^\mathcal{T}$, for $k=1, \cdots, N$. Let $\pmb{\eta}^C_i \in \mathbb{R}^{(N+1) n^{\mathcal{T}}}$ be the vector in which $N+1$ copies of an eigenvector $\pmb{\phi}^\mathcal{T}_i$ are stacked. Then, from \cref{eq:coupled_laplacian} and \cref{eq:uncoupled} we have,
\begin{equation}
\label{eq:eta}
\pmb{L}^C \: \pmb{\eta}^C_i - \lambda^{\mathcal{T}}_i \: \pmb{B}^C \: \pmb{\eta}^C_i = \pmb{L}^+ \: \pmb{\eta}^C_i  
\end{equation}
where non-zero rows in the right hand side term are related to cross-connections between a vertex of $\mathcal{G}^\mathcal{T}$ and its copies on the references, at the same location. Hence, the weight function is equal to one, for each of the cross-connections, and
\begin{equation}
\forall j \in F^\mathcal{T} \quad (\pmb{L}^+ \: \pmb{\eta}^C_i)_j = \sum_{k=1}^N   \phi^\mathcal{T}_{ji} - \phi^{S_k}_{ji}.
\end{equation}
Since $\pmb{\phi}^{S_k}_{i} = \pmb{\phi}^\mathcal{T}_i$ for $\mathcal{G}^{S_k} = \mathcal{G}^\mathcal{T}$, the right hand side term of \cref{eq:eta} is equal to zero and $(\pmb{\eta}^C_i)_{i=1,\cdots, n^\mathcal{T}}$ are eigenvectors of the Coupled Laplacian containing $N+1$ copies of a target graph. 

In addition to the above property, when target graph has eigenvalues with multiplicity higher than one, the coupled term $\pmb{L}^+ \: \pmb{\phi}^C_i = 0$ enforces the shape matching of eigenvectors that do not belong to the kernel of the matrix $\pmb{L}^+$.  

\textbf{Penalization Term.} The computation of matching vectors can be penalized in the initial eigenvectors by incorporating a coefficient $\alpha > 1$ into the Laplacian matrix for the coupled graph, as follows:
\begin{equation}
\pmb{L}^{C\alpha} = \pmb{L}^U + \alpha \: \pmb{L}^+
\end{equation}
The larger $\alpha$, the higher the eigenvalues of non matching eigenvectors such that $\| \pmb{L}^+ \:\pmb{\phi}^C_i \| > 0$.

\textbf{Modal Length.} Let $\pmb{\phi}_k$ be the $k$-th eigenvector of a single graph $\mathcal{G}(\mathcal{V}, \mathcal{E})$, its gradient on the graph can be expressed as
\begin{equation}
    ||\nabla \pmb{\phi}_k ||^2_\mathcal{G} = \sum_{(i, j) \in \mathcal{E}} \left( \frac{\phi_{ik} - \phi_{jk}}{d(\pmb{x}_i, \pmb{x}_j)}\right)^2
\end{equation}
where the sum is over the set of edges $\mathcal{E}$ and $d(\cdot, \cdot)$ is the Euclidean distance. We define the modal length as the ratio between the norm of the eigenvector and its gradient
\begin{equation}
\label{eq:modal_length}
L_k = \frac{||\pmb{\phi}_k||}{||\nabla \pmb{\phi}_k ||_\mathcal{G}}
\end{equation}
with $L_0 = +\infty$, because $\pmb{\phi}_0$ is constant, and $L_k > L_{k+1}$ for $k\neq0$. Note that the unit of $L_k$ is the same as the unit given by the Euclidean distance of points in the graph. Therefore, the modal length can give an insight about the number of eigenmpas, $m$, to produce with the Coupled Laplacian, based on the extent of surface differences we want to detect. Intuitively, representing differences at small scales is easier than at large ones. Therefore, the more modes are used, the larger the range of defect sizes that can be represented.

\begin{algorithm}[t]
    \SetAlgoLined
    \KwData{Source surface and side - $\left(\mathcal{V}^S, \mathrm{S}^{S} \right)$}
    \myinput{Target surface - $\mathcal{V}^\mathcal{T}$}
    \myinput{Neighbours - $k$}
    \myinput{Cross-Fraction - $l$}
    \myinput{Eigenmaps - $m$}
    \KwResult{Target Side - $\mathrm{S}^\mathcal{T}$}
    $\pmb{W}^S \gets \text{Adjacency}(\mathcal{V}^S, k)$\;
    
    $\pmb{W}^\mathcal{T} \gets \text{Adjacency}(\mathcal{V}^\mathcal{T}, k)$\;
    
    $L^S,  L^\mathcal{T}\gets $ Bone lengths from Fiedler vectors\;

    $\alpha \gets L^S / L^\mathcal{T}$ \Comment*[r]{Scale factor}

    $\mathcal{V}^\mathcal{T} \gets \alpha \times \mathcal{V}^\mathcal{T} $\;

    $\mathcal{V}^S_M \gets$ Mirror Source using PCA\;

    $\mathcal{V}^S, \mathcal{V}^S_M\gets$ RANSAC registrations to $\mathcal{V}^\mathcal{T}$\;

    Select target cross-connections set $F^\mathcal{T} \subset \mathcal{V}^\mathcal{T}$ 
    
    Find sets $F^\mathcal{S} \subset \mathcal{V}^S$ and $F^\mathcal{S}_M \subset \mathcal{V}^S_M$ as in \cref{eq:nearest}

    $L^C\gets\text{Coupled Laplacin}(W^\mathcal{T}, W^S, F^\mathcal{T}, F^\mathcal{S}, F^\mathcal{S}_M)$\;

    $\{\lambda_i^C\}_{i=0}^m$, $\{\pmb{\phi}_i^C\}_{i=0}^m\gets$ Eig$(L^C, m)$\;

    $\pmb{\Phi}^\mathcal{T}, \pmb{\Phi}^S, \pmb{\Phi}^S_M \gets$ Split($\{\pmb{\phi}_i^C\}_{i=0}^m$)\;

    QR decompose $\pmb{\Phi}^\mathcal{T}(F^\mathcal{T}, :), \pmb{\Phi}^S(F^S, :), \pmb{\Phi}^S_M(F^S_M, :)$\;

    \uIf{$d_G(\pmb{Q}^\mathcal{T}, \pmb{Q}^{S}) \le d_G(\pmb{Q}^\mathcal{T}, \pmb{Q}^{S}_M)$}{$\mathrm{S}^\mathcal{T}$ = $\mathrm{S}^S$ \Comment*[r]{Same side}}
    \Else{$\mathrm{S}^\mathcal{T}$ = $\neg \mathrm{S}^S$ \Comment*[r]{Opposite side}}
    
    \caption{Spectral-Based Bone Side Estimation (BSE) using aligned embeddings from the Coupled Laplacian.}
    \label{alg:BSE}
\end{algorithm}

\section{Spectral-Based BSE}
\label{sec:SB_BSE}

In \cref{alg:BSE} we provide full details on the spectral-based algorithm for BSE introduced in \cref{sec:settings}. It takes as input the set of points, i.e. graph vertices, representing source and target surfaces , $\mathcal{V}^S$ and $\mathcal{V}^\mathcal{T}$, respectively, and the target side, $\mathrm{S}^{S}$. Additional input parameters are the number of neighbours used to create the $k$-NN graphs from the point clouds, the fraction of target nodes used to create cross-connection, $l$, and the dimension of the embeddings generated through the Coupled Laplacian, $m$. 

As first step, the single adjacency matrices are built from the set of vertices using an RBF kernel, as in \cref{eq:RBF}. After that, the length of each shape is computed only with the information carried by the Fiedler vector, $f$, as follows
\begin{equation}
    L = || x_{\text{max}} - x_{\text{min}}||^2
\end{equation}
where $x_{\text{max}} = x_{\argmax{f}}$ and $x_{\text{min}} = x_{\argmin{f}}$ are the points corresponding to the maximum and minimum values of the Fiedler vector, respectively. In this way, the input surfaces do not need to lie on the same Euclidean frame in order to have comparable length measurements. Even if Fiedler vectors of different shapes may have opposite sign, the inversion of maximum and minimum does not affect the bone length calculation. Note that, as extension for a more precise length computation, it is also possible to select the $M$ maximum and minimum points of the Fiedler vector and compute $x_{\text{max}}$ and $x_{\text{min}}$ as the barycenters of those points. The ratio between source and target lengths gives a scaling factor, $\alpha$, that we use to scale the target surface in order to match the length of the reference. The mirrored version of the source bone, $\mathcal{V}^S_M$, representing its contralateral, is then generated by flipping it around the point cloud second principal component and both versions of the reference are registered to the target with RANSAC algorithm \cite{fischler_random_1981}. Interestingly, the adjacency matrix of the two references, before and after registration, is the same as the original one and, therefore, we can avoid its computation multiple times. 

\begin{figure}
    \centering
    \includegraphics[width=0.37\textwidth]{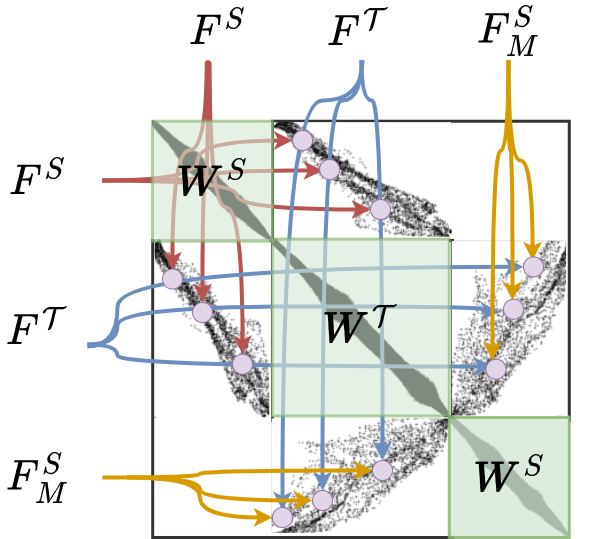}
    \caption{Graphical representation of the structure of the coupled adjecency matrix. $\pmb{W}^{S}$ and $\pmb{W}^\mathcal{T}$ are the source and target adjacency matrices, while $F^\mathcal{T}$, $F^S$ and $F^S_M$ are the set of nodes for cross-connections of target, source and mirrored source. Note that the nodes are sorted in groups such that the eigenmaps derived from the Coupled Laplacian can be easily split into the single components.}
    \label{fig:coupled_graph}
\end{figure}
\begin{figure*}
    \centering
    \includegraphics[width=0.9\textwidth]{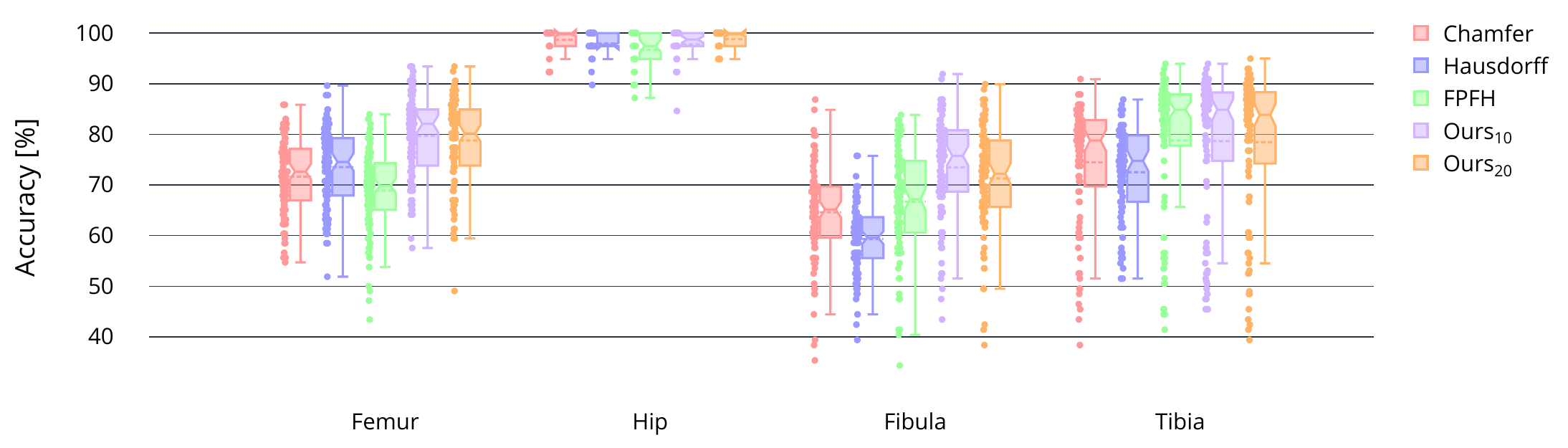}
    \caption{Box plot of the accuracy of various methods for BSE depending on the source shape. Each measurement is the accuracy obtained using a different source shape among the ones available in the proposed benchmark. The median is highlighted with the notch while the mean is represented by the dashed line. }
    \label{fig:BSE_robustness}
\end{figure*}

After the described pre-processing, the core of the algorithm based on the Coupled Laplacian takes place. The sub-set of target nodes used for cross-connections is stochastically selected and the corresponding points on the two versions of the reference are queried as in \cref{eq:nearest}. Alternatively, assuming not flat surfaces, other methods for cross-connections can be used instead of NN, e.g. spectral clustering medoids. Hence, the three graphs can now be connected and the Coupled Laplacian computed from the whole weighted adjacency matrix. For more clarity, the structure of the coupled graph adjacency matrix is depicted in \cref{fig:coupled_graph}. Therefore, the first $m$ coupled eigenmaps are computed with a given eigensolver and split into the components of each of the three single shapes, as in \cref{eq:split}. Finally, the Grassman distances, $d_G(\cdot, \cdot)$, of the QR normalized aligned eigenmaps restricted to the cross-connection sets are compared in order to predict the target side. The latter is equal to $\mathrm{S}^{S}$ if the Grassman distance between target and reference is lower than the distance between target and mirrored reference, otherwise it will be the opposite body side. 

\section{Robustness to Source Variation}
\label{sec:robustness}
In this section we briefly discuss the robustness with respect to the selected source shape of the methods tested in \cref{sec:results}. 
\begin{table}[t]
\footnotesize
\caption{Minimum and maximum accuracy of human BSE for each bone structures using different sources. All the matching methods are applied after RANSAC registration. The overall best performing methods are highlighted in boldface.}
    \centering
\begin{tabularx}{0.45\textwidth}{ll|YYYY}
\toprule
 & Method & Femur & Hip & Fibula & Tibia \\
 \midrule  
  
 \multirow{5}{*}{\rotatebox[origin=c]{90}{\textbf{Best}}} & Hausdorff & 89.62 & \textbf{100.0} & 75.76 & 86.87\\
  
 & Chamfer & 85.85 & \textbf{100.0} & 86.87 & 90.91 \\

 & FPFH \cite{rusu_fast_2009} & 83.96 & \textbf{100.0} & 83.84 & 93.94\\ %

 & $\textbf{Ours}_{20}$ & \textbf{93.40} & \textbf{100.0} & 89.90 & \textbf{94.95} \\

 & $\textbf{Ours}_{10}$ & \textbf{93.40} & \textbf{100.0} & \textbf{91.92} & 93.94 \\

 \midrule  
  
 \multirow{5}{*}{\rotatebox[origin=c]{90}{\textbf{Worst}}} & Hausdorff & 51.89 & 89.74 & 39.39 & \textbf{51.51}\\
  
 & Chamfer & 54.72 & 92.31 & 35.35 & 38.38 \\

 & FPFH \cite{rusu_fast_2009} & 34.34 & 87.18 & 34.34 & 41.41 \\ %

 & $\textbf{Ours}_{20}$ & 49.06 & \textbf{94.87} & 38.38 & 39.39 \\

 & $\textbf{Ours}_{10}$ & \textbf{57.55} & 84.62 & \textbf{43.43} & 45.45 \\
  
\bottomrule
\end{tabularx}
    \label{tab:benchmark_results_min_max}
\end{table}

\textbf{Bone Side Estimation.} \cref{fig:BSE_robustness} shows the variation of the BSE accuracy, with respect to the source bone, for different bones and different methods. We recall that in each experiment one bone is chosen as source and the side is inferred on all the other bones of the same category. Therefore, we automatically have $N_b$ accuracy measurements for each class, where $N_b$ is the number of samples of the bone class $b$. The average accuracy of these experiments is reported in \cref{tab:benchmark_results} of the main manuscript. While median and average accuracy are generally higher using our technique in all the bones, the values spread is similar between methods. Additionally, we report in \cref{tab:benchmark_results_min_max} the maximum and the minimum accuracies obtained using different sources. We can observe that the maximum accuracy is always achieved with our method, both using $10$ or $20$ eigenmaps for matching. Moreover, it is also better in the worst case for all the bones classes except for the tibia, for which the Hausdorff distance seems to be better. Nevertheless, as also shown in \cref{fig:BSE_robustness}, there is no much difference in the span between best and worst cases among different methods. This suggests that, to improve the performance of our BSE technique, and also of other methods, a proper choice of source bone is essential.

\begin{figure}
    \centering
    \includegraphics[width=0.45\textwidth]{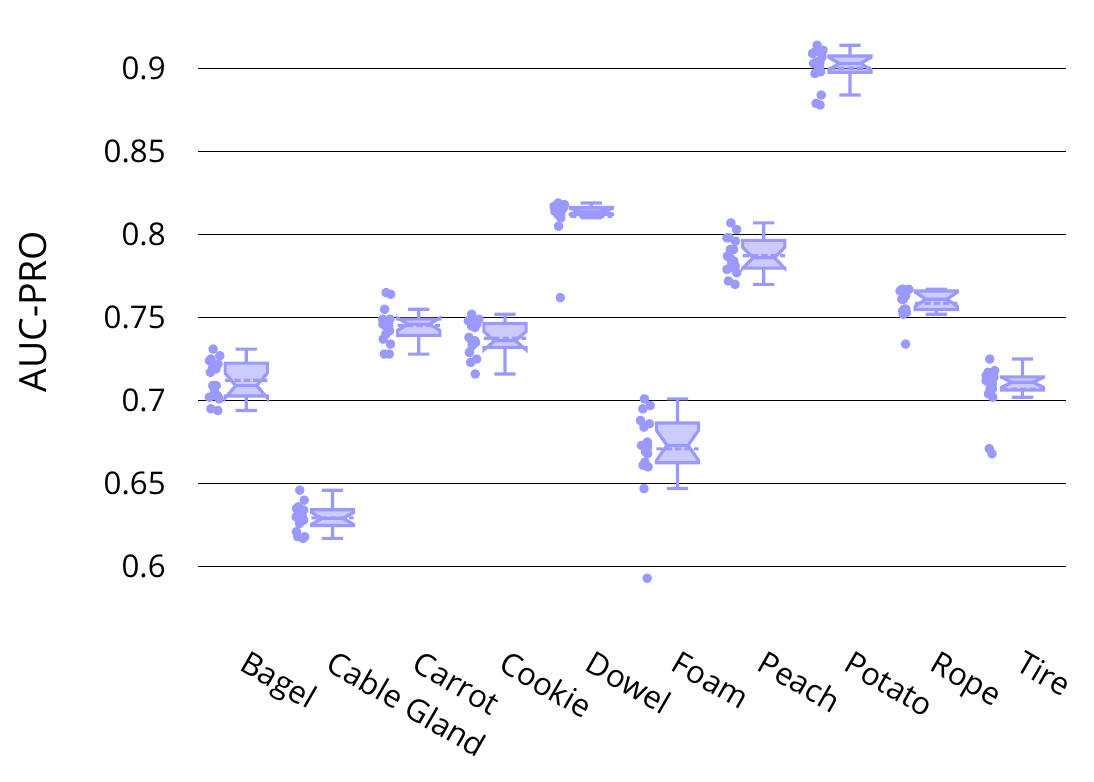}
    \caption{Area under the PRO curve (AUC-PRO) with an upper integration limit of 0.3 for each category of the MVTec 3D-AD dataset. Each score, represented by a dot, is computed with a different source shape selected from the train set. The median is highlighted with the notch while the mean is represented by the dashed line. }
    \label{fig:robustness_ad}
\end{figure}
\begin{figure*}
    \centering
    \includegraphics[width=\textwidth]{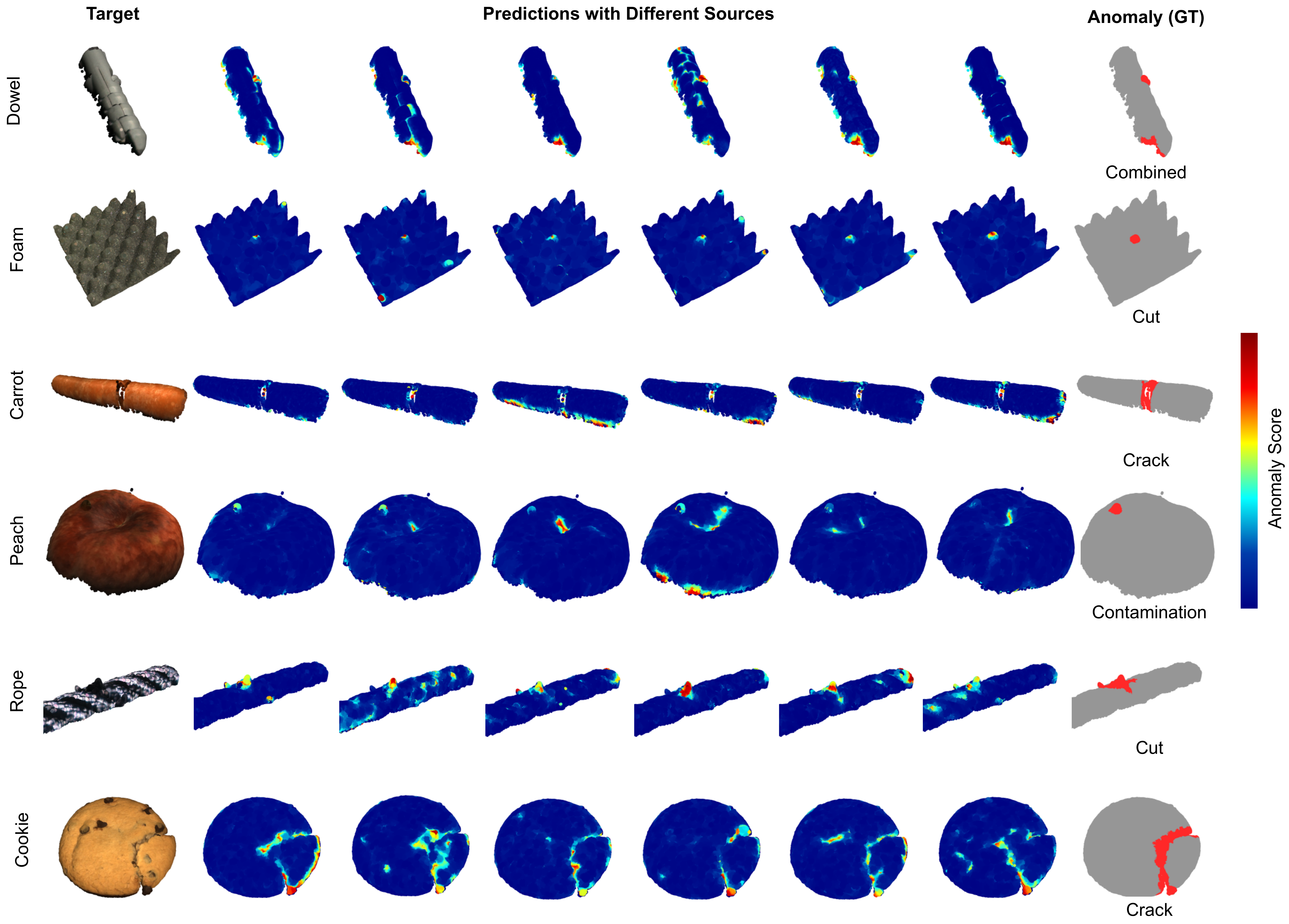}
    \caption{Graphical comparison of 3D anomaly localization using different source shapes on different objects. The target example with RGB colors and the GT anomaly are shown on the left and right, respectively. In the middle $6$ different results obtained changing the anomaly-free source are reported. }
    \label{fig:robustness_ad_graphical}
\end{figure*}
\textbf{Anomaly Localization.} We tested our method on the 3D anomaly localization task using $20$ different anomaly-free sources randomly drawn from the train set of the MVTec 3D-AD dataset. For each train sample, and for each category, we computed the normalized area under the PRO curve with an upper integration limit of 0.3 and their distribution is shown in \cref{fig:robustness_ad}. All the experiments are performed using the Coupled Laplacian with $m=200$ and $l=1$. We can observe that the results are much more stable and less spread than in the BSE task. This is likely due to the fact that different samples of the same industrially manufactured product are very similar and, therefore, all equally good as source shape. On the other hand, bones may vary in size and slightly in form from patient to patient. Hence, the selection of a good representative source is essential to enhance the capabilities of the proposed matching technique. 

Moreover, in \cref{fig:robustness_ad_graphical} we compare graphically the results obtained on some test samples using different anomaly-free sources. We can observe the strength of the proposed method which always correctly localizes the anomalies, even with different sources. In opposition, some false positives are sometimes present, especially on the borders of the shapes. This weakness is due to fact that the considered point clouds are not closed surfaces, since they are acquired with objects placed on a flat surface. Therefore, the points on the edges between object and surface are likely to contain noise which may produce different coupled embeddings leading to a detection of local surface differences. Interestingly, in the peach sample, some anomalies are located in the area of the steam end because only some anomaly-free sources has it. The peach test example shown in \cref{fig:robustness_ad_graphical} does not have the steam and therefore, when compared with a source that has it, a surface difference is highlighted in that specific area. 

\begin{figure*}
    \centering
    \includegraphics[width=0.7\textwidth]{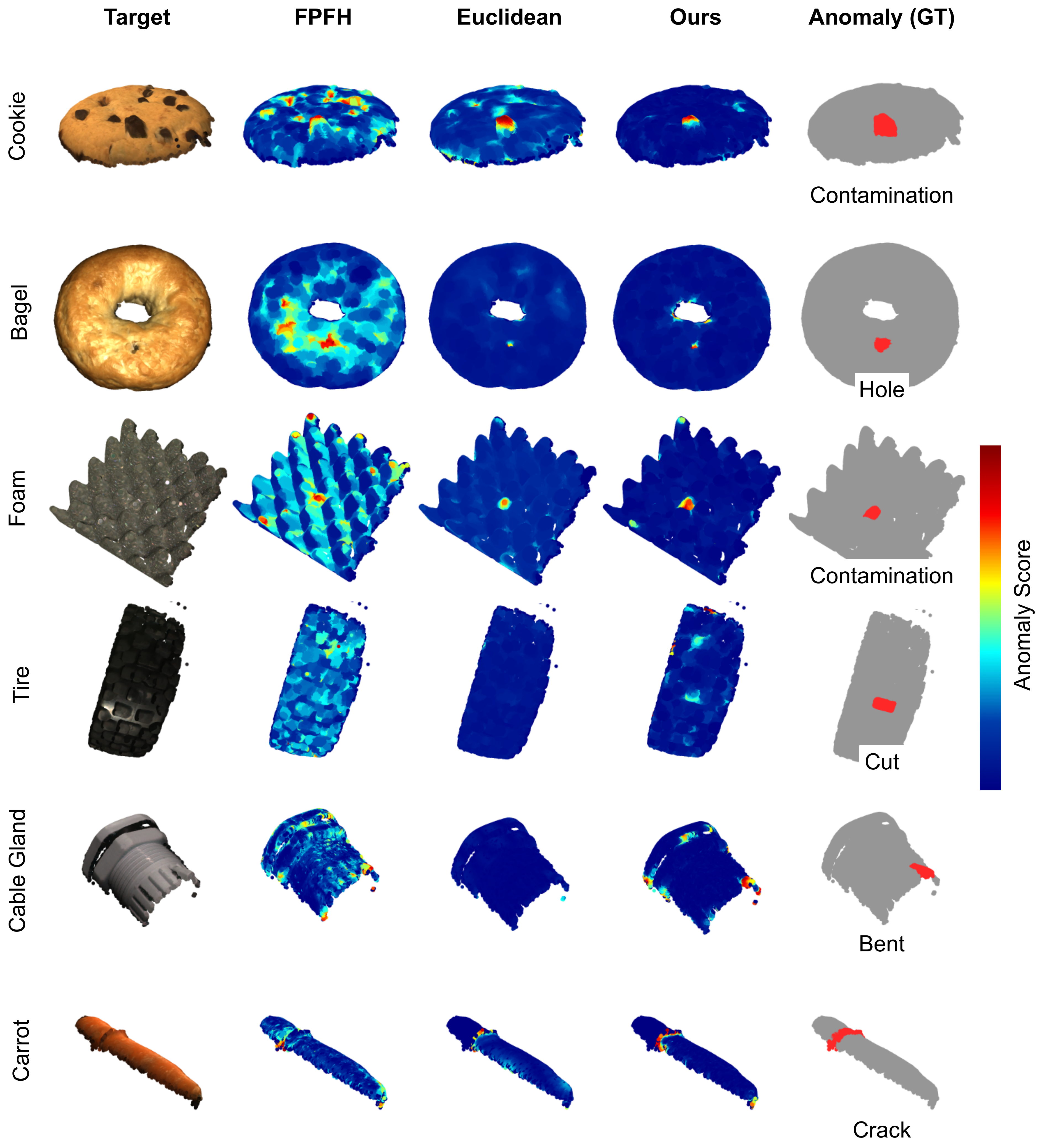}
    \caption{Graphical comparison of 3D anomaly localization using different methods on different objects. The target example with RGB colors and the GT anomaly are shown on the left and right, respectively. In the middle the results obtained with FPFH, Euclidean distance and distance between aligned spectral spaces are reported. }
    \label{fig:ad_graphical_comp_supp}
\end{figure*}
\section{Graphical Comparison}
In this section we first provide more graphical comparisons of different methods on the anomaly localization task. After that, we show the difference between eigenmodes produced with the Coupled Laplacian with respect to the ones generate independently on the single geometries. 

In \cref{fig:ad_graphical_comp_supp} the anomaly scores obtained using FPFH features comparison \cite{rusu_fast_2009}, Euclidean distance and Coupled Laplacian, with $m=200$ and $l=1$, on different samples are compared. For a fair comparison, all the techniques are performed with the same anomaly-free source sample and after affine CPD \cite{myronenko_point-set_2010} source to target registration. We observed that FPFH tends to overestimate anomalies, whereas the Euclidean distance underestimates them, leading to misleading detections. In contrast, our method achieves a better trade-off and improves the localization of anomalies. With reference to the modal length of \cref{eq:modal_length}, decreasing the number of modes used for spectral comparison we will obtain similar results to the Euclidean method, while increasing it, the localization will tend to the one of FPFH. Once again, the parameter $m$ plays a crucial role in our method, helping us select surface differences that are tailored to the specific task at hand.  

\begin{figure*}
    \centering
    \includegraphics[width=0.8\textwidth]{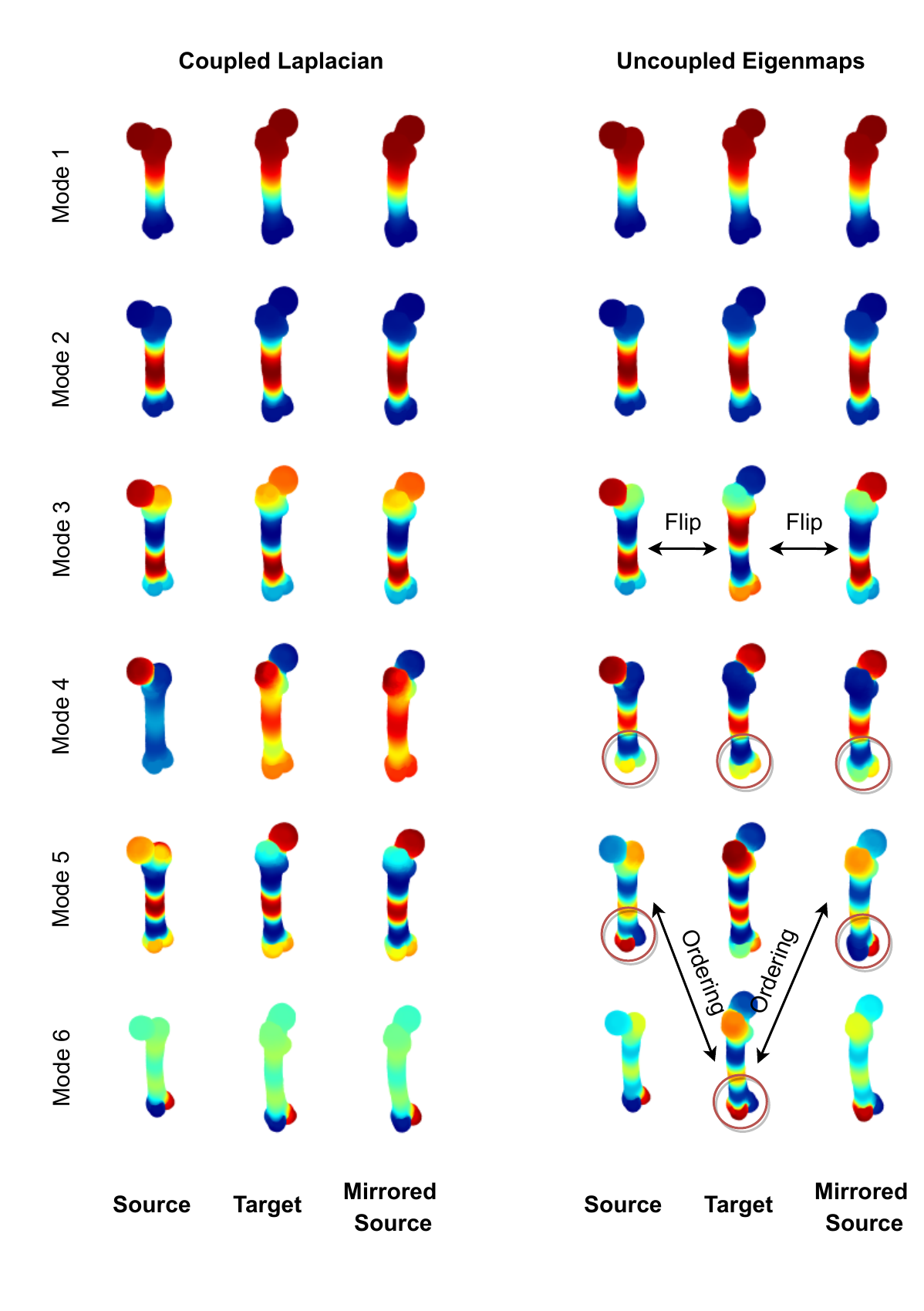}
    \caption{Graphical comparison of the eigenmaps obtained through Coupled Laplacian and with independent eigendecomposition of point cloud graphs, i.e. uncoupled eigenmaps. The first $6$ eigenmodes are shown in both cases
 for source, target and mirrored target in the case of femur bones. }
    \label{fig:coupled_embeddings}
\end{figure*}

\begin{figure*}
    \centering
    \includegraphics[width=\textwidth]{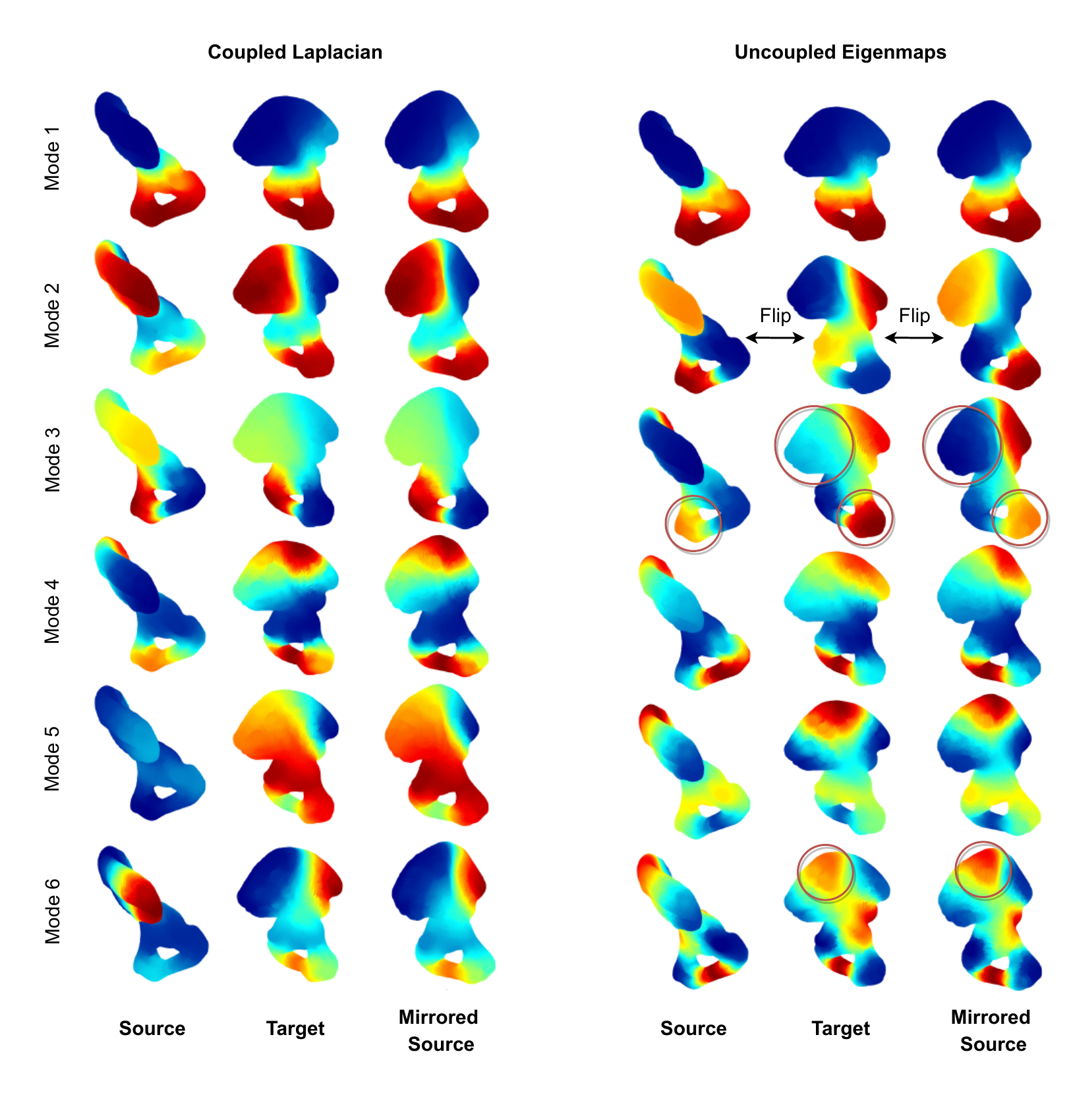}
    \caption{Graphical comparison of the eigenmaps obtained through Coupled Laplacian and with independent eigendecomposition of point cloud graphs, i.e. uncoupled eigenmaps. The first $6$ eigenmodes are shown in both cases for source, target and mirrored target in the case of hip bones.}
    \label{fig:coupled_embeddings2}
\end{figure*}

\begin{figure*}
    \centering
    \includegraphics[width=\textwidth]{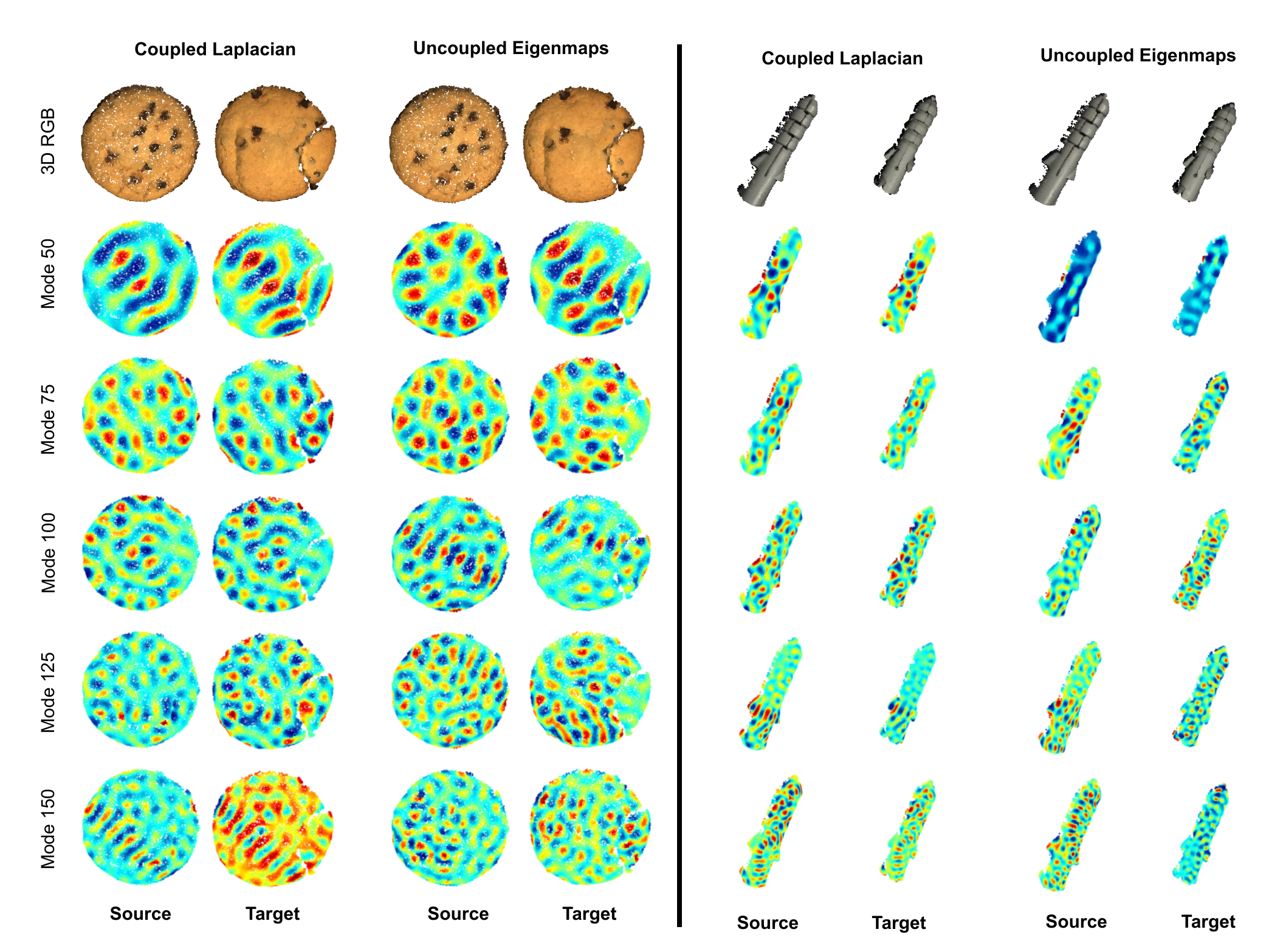}
    \caption{Graphical comparison of the eigenmaps obtained through Coupled Laplacian and with independent eigendecomposition of point cloud graphs, i.e. uncoupled eigenmaps. Some eigenmodes of increasing number are shown for two samples of the MVTec 3D-AD dataset.}
    \label{fig:coupled_embeddings3}
\end{figure*}

In \cref{fig:coupled_embeddings} and \cref{fig:coupled_embeddings2} the first  $6$ eigenmodes of a femur and hip BSE problem produced with the Coupled Laplacian are shown and compared with the ones obtained thought independent eigendecomposition of target and source, i.e. uncoupled eigenmaps. In both cases the target side is opposite to the source and therefore the best matching coupled eigenmaps are with the mirrored source. This is because its RANSAC registration to the target shape is more precise and so the added cross-connections. On the contrary, the cross-edges between source and target are weaker leading to less precise coupled modes and larger Grassmann distance, which result in a correct BSE. In the uncoupled eigenmaps we can instead observe both the eigenvalue ordering and sing disambiguity of the eigendecomposition. In the femur example, eigenmode $3$ has flipped sign in source and target, while eigenvalues $5$ and $6$ are inverted. Moreover, we can observe slight local variations of modes in the condyles area of the distal femur (highlithed with red circles). Similar observations can be also derived from the uncoupled modes of the hip bones in \cref{fig:coupled_embeddings2}.

A comparison of higher modes is instead shown for two samples of the anomaly detection task in \cref{fig:coupled_embeddings3}. In relation to the modal length defined in \cref{eq:modal_length}, it is evident that lower modes represent larger scales compared to higher modes, enabling them to better capture differences. Moreover, the proposed coupling process leads to aligned eigenmaps for source and target shapes. Uncoupled eigenmodes exhibit completely different patterns, rendering them impractical for matching purposes without a reordering process, which complexity increases as the represented scale decreases. On the other hand, modes derived from the Coupled Laplacian exhibit the exact same pattern on both source and target point clouds, making them feasible for direct comparison. However, it is worth noting that as we increase the mode number, there might be an observable weakening in the coupling of the intensities of the eigenvectors, despite the preservation of the underlying patterns. In the computation of similarity scores using aligned embeddings, the observed coupling weakening with increasing mode numbers may not significantly impact the results. In fact, this characteristic could be leveraged to our advantage when determining the truncation number for the eigenspaces, highlighting the adaptability and effectiveness of our method.

\begin{table}[t]
\footnotesize
\caption{Accuracy of cross-species BSE for femur and tibia using different methods. All the matching methods are applied after RANSAC registration. \textbf{S} and \textbf{H} stands for Sheep and Human, respectively, and the overall best performing methods are highlighted in boldface.}
    \centering
\begin{tabularx}{0.45\textwidth}{ll|YY|Y}
\toprule
 & Method & Femur & Tibia & Mean\\
 \midrule  
  
 \multirow{5}{*}{\rotatebox[origin=c]{90}{\textbf{S $\rightarrow$ S}}} & Hausdorff & 94.37 & 97.14 & 90.48\\
  
 & Chamfer & 94.29 & 88.57 & 94.88\\

 & FPFH \cite{rusu_fast_2009} & 97.14 & 97.14 & 97.14 \\ %

 & $\textbf{Ours}_{20}$ & \textbf{97.30} & \textbf{97.30} & \textbf{97.30} \\

 & $\textbf{Ours}_{10}$ & \textbf{97.30} & \textbf{97.30} & \textbf{97.30} \\

 \midrule  
  
 \multirow{5}{*}{\rotatebox[origin=c]{90}{\textbf{H $\rightarrow$ S}}} & Hausdorff & 72.22 & 69.44 & 70.83\\
  
 & Chamfer & 69.44 & 75.00 & 72.22\\

 & FPFH \cite{rusu_fast_2009} & 77.78 & 75.00 & 76.39\\ %

 & $\textbf{Ours}_{20}$ &  69.44 & 72.22 & 70.83\\

 & $\textbf{Ours}_{10}$ & \textbf{83.33} & \textbf{77.78} & \textbf{80.56}\\

  \midrule  
  
 \multirow{5}{*}{\rotatebox[origin=c]{90}{\textbf{S $\rightarrow$ H}}} & Hausdorff & 57.94 & 63.00 & 60.47\\
  
 & Chamfer & 62.62 & \textbf{66.00} & 64.31\\

 & FPFH \cite{rusu_fast_2009} &  \textbf{63.89} & 63.00 & 63.45\\ %

 & $\textbf{Ours}_{20}$ & 63.55  & 60.00 & 61.78\\

 & $\textbf{Ours}_{10}$ & 63.55 & \textbf{66.00} & \textbf{64.78} \\
  
\bottomrule
\end{tabularx}
    \label{tab:cross}
\end{table}
\begin{figure}
    \centering
    \includegraphics[width=0.45\textwidth]{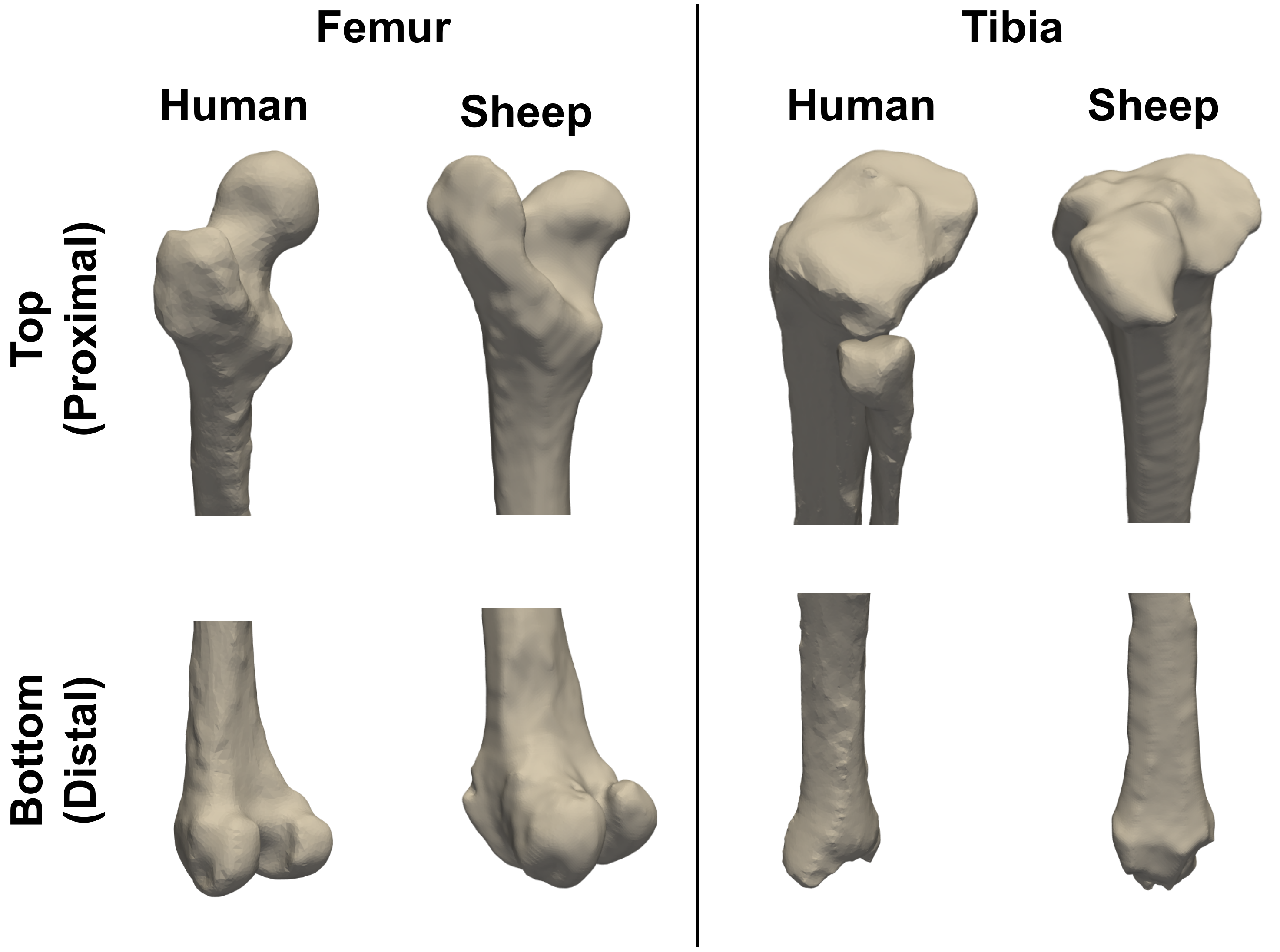}
    \caption{Comparison of human and sheep femur and tibia bones. For both of them, the top and lower parts are shown separately to highlight their differences.}
    \label{fig:human_sheep}
\end{figure}

\section{Cross-Species BSE}
We test here the generalization of the proposed BSE algorithm on an internal dataset of sheep femur and tibia bones and we discuss its cross-species capabilities, i.e. when source and target bones are from different species. The latter poses an interesting challenge in the medical field, as tests are often performed on animals before being expanded to humans. The experiments only on sheep bones are performed as described in \cref{sec:settings} for the public human benchmark. While, for the cross-species experiments, each bone of one species is chosen once as source to infer the side of the bones of the same category from the other species. The results of these sets of experiments are reported in \cref{tab:cross}. We can observe that, the sheep-to-sheep BSE is much more effective then the human-to-human comparison reported in the main manuscript, achieving almost correct estimation in all the sample. This is due to the fact that sheep femur and tibia have less symmetric shapes than their human counterparts, hence making easier the side estimation. Moreover, all the samples in our internal datasets are acquired with the same settings and therefore there is small sample variability, unlike the human benchmark we proposed to collect.  

However, despite our method maintaining its superiority, performance drops are noted for cross-species tasks. A clearer understanding of this phenomenon can be obtained from \cref{fig:human_sheep}, where human and sheep femur and tibia bones are graphically compared. While the overall shapes are similar, some major differences can be noticed between species. For instance, the sheep proximal femur is similar to the human distal part. Such variations, and other artifacts, may affect the result of the initial registration, leading to wrong cross-connections and miss-aligned eigenmaps. Furthermore, while bones of different species are globally similar, local differences introduce complexities in the BSE task. The global similarity score derived from the Coupled Laplacian is therefore affected by these differences, leading to a more challenging task of aligning and comparing bones across species. Addressing this challenge will contribute to advancing the field of BSE and refining the understanding of global similarities and localized distinctions in diverse bone structures.

\section{Scalability}
\begin{figure}
    \centering
    \includegraphics[width=0.45\textwidth]{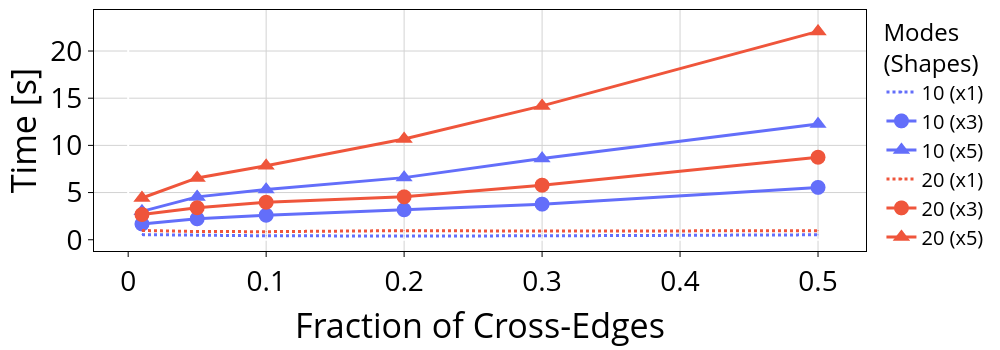}
    \caption{Average computational time of the eigendecomposition using different combinations of number of shapes ($1$, $3$ or $5$) and modes ($10$ or $20$).}
    \label{fig:scalability}
\end{figure}
\begin{figure}
    \centering
    \includegraphics[width=0.45\textwidth]{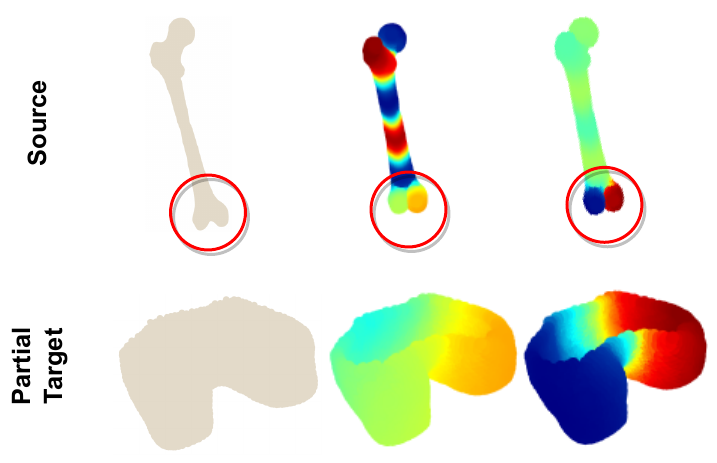}
    \caption{Coupled modes correctly aligned in a partial matching scenario.}
    \label{fig:partial_matching}
\end{figure}
The efficiency trade-off of decomposing larger Laplacians is illustrated in \cref{fig:scalability}. The average computational time is reported for the eigendecomposition, using $10$ and $20$ maps, coupling the same shape with itself $3$ or $5$ times. As expected, larger, i.e. number of shapes, and denser, i.e. number of cross-connections, is the Laplacian, higher is the time needed for the computation. Scalability might be a concern in tasks like shape retrieval with numerous classes. However, simple measures, e.g. downsampling shapes and interpolating eigenmaps, can reduce the complexity. Given the intrinsic coupling of modes, faster, but randomized, solvers, e.g. AMG, are still precise and have been used, cutting the need of resources.

\section{Partial Matching}
In \cref{fig:partial_matching} we show a graphical example of coupled eigenmaps derived from partially matching shapes. More in detail, an entire femur bone is matched with the distal part, i.e. lower part, of another femur. In this case, artificial cross-connections are added only on the partial match of the source shapes, leading to correctly aligned eigenmaps. Namely, the embeddings are successfully coupled in the distal area of the femur while they are independent on the remaining surface of the source. This example shows the potential of the Coupled Laplacian operator in different, and more complex, scenarios. 

\end{document}